\documentclass[10pt,twocolumn,letterpaper,dvipsnames,table]{article}

\usepackage{cvpr}
\usepackage{adjustbox}
\usepackage{multirow}
\usepackage{float}
\usepackage{placeins}
\usepackage{amsthm}
\usepackage{wrapfig}
\usepackage{xcolor}

\definecolor{cvprblue}{rgb}{0.21,0.49,0.74}
\usepackage[pagebackref,breaklinks,colorlinks,allcolors=cvprblue]{hyperref}

\def\paperID{42989} 
\def\confName{CVPR}

\title{Landscape-Awareness for Geometric View Diffusion Model}

\author{
Yan-Ting Chen$^{*}$, Hao-Wei Chen$^{*}$, Tsu-Ching Hsiao, and Chun-Yi Lee\\
Elsa Lab, National Taiwan University\\
\tt{\small{r13922027@g.ntu.edu.tw, d13922023@csie.ntu.edu.tw,}}\\
\tt{\small{joehsiao.x@gmail.com, cylee@csie.ntu.edu.tw}}\\
{\small $^{*}$Equal contribution}
}

\begin{document}
\maketitle

\begin{abstract}
Accurate camera viewpoint estimation under sparse-view conditions remains challenging, particularly in two-view scenarios. Recent approaches leverage diffusion models such as Zero123 to synthesize novel views conditioned on relative viewpoint, showing promising results when repurposed for viewpoint estimation via optimization with MSE loss. However, existing methods often suffer from non-convex loss landscape with numerous local minima, making them sensitive to initialization and reliant on na\"ive multi-start strategies. We analyze these optimization challenges and visualize failure cases, showing that geometric ambiguities, such as symmetry and self-similarity, can mislead gradient-based updates toward incorrect viewpoints. To address these limitations, we propose a score-based method that reshapes the optimization landscape to guide updates toward the ground-truth viewpoint, followed by a refinement stage using a viewpoint-conditioned diffusion model. Experiments show that our method improves convergence, reduces reliance on brute-force sampling, and achieves competitive accuracy with higher sample-efficiency.
\end{abstract}

\vspace{-0.5em}
\section{Introduction}
\vspace{-0.5em}
\label{sec:introduction}

Camera pose estimation constitutes a fundamental component in a wide range of applications, including robotics~\citep{DBLP:conf/iros/YuCPP19}, structure from motion~\citep{DBLP:conf/cvpr/SchonbergerF16, DBLP:conf/eccv/WeiZLFX20}, visual SLAM~\citep{DBLP:journals/trob/Mur-ArtalMT15, DBLP:conf/eccv/EngelSC14}, augmented reality and virtual reality~\citep{belghit2018vision, DBLP:journals/frvir/PantelerisMA21, DBLP:conf/cvpr/YuYTK21}, and 3D reconstruction~\citep{jiang2024few, DBLP:conf/eccv/MildenhallSTBRN20,DBLP:conf/cvpr/ZhouT23}. Traditional methods typically establish pose estimation through feature correspondence, using either hand-crafted features~\citep{DBLP:journals/pami/LagunaM23, DBLP:conf/eccv/BayTG06, DBLP:journals/ijcv/Lowe04} or learned features~\citep{detone2018superpoint, revaud2019r2d2} to extract and match keypoints across images~\citep{DBLP:conf/cvpr/SarlinDMR20, DBLP:conf/eccv/ChenLZTZFMTQ22}. While feature-based approaches demonstrate strong performance in dense-view settings with sufficient overlap, they frequently fail in sparse-view scenarios where substantial viewpoint differences result in unreliable correspondences. To address these limitations, many recent works~\citep{DBLP:conf/cvpr/Sinha0TGL23, DBLP:conf/eccv/ZhangRT22, DBLP:conf/3dim/LinZRT24, DBLP:conf/cvpr/YuYTK21, DBLP:conf/cvpr/NiemeyerBMS0R22, tang2024aden} have directed their attention toward sparse-view scenarios. These methods diverge from the conventional pipeline and adopt data-driven techniques to learn geometric priors of objects. Nevertheless, contemporary approaches continue to face substantial challenges in highly sparse two-view settings. In such scenarios, large viewpoint differences lead to minimal feature overlap and introduce geometric ambiguities on occluded sides of objects, which compromise the reliability of pose estimation.

Since feature correspondence-based approaches often fail in sparse-view scenarios due to unreliable matches, recent methods leverage diffusion models~\citep{DBLP:conf/nips/HoJA20, DBLP:conf/nips/SongE19, DBLP:conf/iclr/0011SKKEP21}, which can model complex image distributions while conditioning on modalities~\citep{zhan2024conditional}, such as text~\citep{saharia2022photorealistic, zhang2023adding, kawar2023imagic}, mask~\citep{couairon2022diffedit}, and pose~\citep{DBLP:conf/iccv/LiuWHTZV23, shi2023zero123++}. A representative example is Zero123~\citep{DBLP:conf/iccv/LiuWHTZV23}, which generates novel views from a reference image given a target relative pose parameterized in spherical coordinates. Its pose-conditioned generation has enabled applications such as novel view synthesis, image-to-3D generation~\citep{liu2023syncdreamer, liu2023one, DBLP:conf/iccv/LiuWHTZV23, wu2023ifusion, zhao2024sparse}, and pose estimation~\citep{cheng2023id, wu2023ifusion, zhao2024sparse}. Building on this, methods like ID-Pose~\citep{cheng2023id} and iFusion~\citep{wu2023ifusion} reformulate pose estimation as an inverse problem by optimizing the pose via mean squared error (MSE) in the diffusion noise space. Given a reference and query image, they first compute the MSE in the noise space and backpropagate gradients with respect to the conditioned pose. Subsequently, they apply gradient descent to optimize the estimated pose such that the generated view closely aligns with the query image. These approaches demonstrate strong performance even under large viewpoint differences due to the robust generative capabilities of the underlying diffusion models.

\begin{figure*}[t]
  \vspace{-2.5em}
  \centering
  \includegraphics[width=1.0\linewidth]{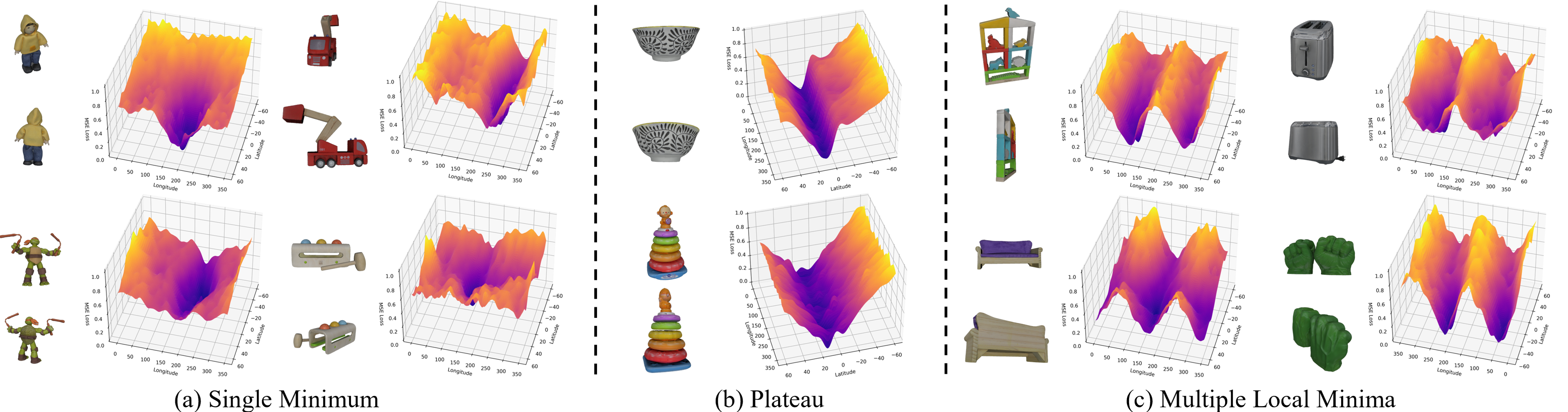}
  \vspace{-2.0em}
  \caption{\textbf{3D MSE Loss Landscape.} Each object is associated with two views, serving as the reference image and the query image. The coordinates follow a spherical system, where the x- and y-axes denote latitude and longitude, and the z-axis indicates the normalized MSE magnitude. The generation procedure is detailed in Appendix~\ref{sec:visualization}.
  (a) Some landscapes exhibit a single clear minimum, enabling gradient descent to reach the global optimum easily.
  (b) Some display a plateau along the longitudinal direction, reflecting continuous symmetry.
  (c) Others contain two distinct local minima.
  In the latter two cases, iFusion's optimization process is prone to getting stuck in a local minima, preventing convergence to the optimal solution. Additional visualizations of MSE loss landscapes are provided in Appendix~\ref{sec:visualization}.}
  \label{fig:ifusion_landscape}
  \vspace{-1.5em}
\end{figure*}

Leveraging a pretrained pose-conditioned diffusion model for camera pose estimation offers two advantages. First, it exploits the generative capability of Zero123 to enable extreme two-view estimation. Second, unlike traditional energy-based methods~\citep{DBLP:conf/eccv/ZhangRT22,DBLP:conf/3dim/LinZRT24,murphy2021implicit,DBLP:conf/wacv/HoferKMZ23,DBLP:conf/iccvw/HaugaardHI23}, which rely on brute-force sampling to find the optimal pose, these approaches utilize the MSE loss computed in Zero123's noise space as an energy function, enabling direct gradient-based optimization of the pose. For instance, RelPose~\citep{DBLP:conf/eccv/ZhangRT22} samples up to 50,000 candidate poses, which shows significant computational inefficiency. Approaches training via likelihood maximization often suffer from non-smooth energy landscapes, making gradient-based optimization unstable and requiring extensive sampling. In contrast, leveraging a pretrained pose-conditioned diffusion model as an energy function provides smoother gradients and supports end-to-end optimization.
Despite this improvement, these methods still require multiple initializations to avoid convergence to incorrect viewpoints. 
This suggests that while the Zero123 noise-space MSE smooths the optimization landscape, we conjecture the presence of local minima on the landscape.
\vspace{-0.5em}

\begin{figure*}[t]
  \centering
  \includegraphics[width=1.0\linewidth]{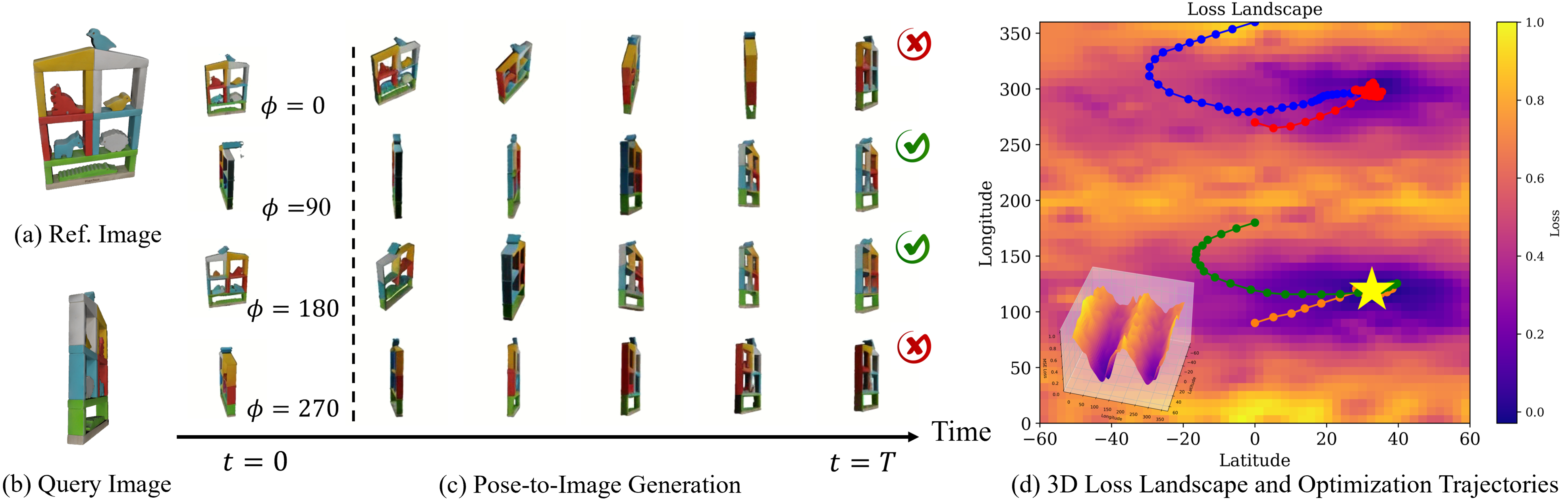}
  \vspace{-1.5em}
  \caption{
  (a) \& (b) Reference and query images of the object, captured from different camera poses.
  (c) Images generated by feeding the poses from optimization trajectory back into Zero123~\citep{DBLP:conf/iccv/LiuWHTZV23}. Although all timestep-\(T\) images appear visually similar to the query, only two accurately reproduce the object's correct appearance—white and blue in the front, yellow and red in the back—indicating correct pose alignment.
  (d) 2D MSE landscape with optimization trajectories initialized from four different starting poses at longitudes \(0^\circ\), \(90^\circ\), \(180^\circ\), and \(270^\circ\). Two of the trajectories converge to local minima. The bottom-left inset shows the 3D landscape for a clearer comparison. }
  \label{fig:trajs}
  \vspace{-1.em}
\end{figure*}

To examine the hypothesis that the optimization landscape contains local minima, Fig.~\ref{fig:ifusion_landscape} provides a visualization of this landscape. Specifically, the conditioned pose in Zero123 is varied while keeping the input image pair fixed and computing the corresponding loss defined in Eq.~(\ref{eq:ifusion_optimization_formula}). The visualizations reveal distinct landscape characteristics across different objects. As shown in Fig.~\ref{fig:ifusion_landscape}~(a), certain objects exhibit smooth surfaces with a single dominant basin, allowing gradient descent to consistently converge to the correct pose. In contrast, other examples in Figs.~\ref{fig:ifusion_landscape}~(b) and (c) display multiple valleys and extended plateaus, indicating the presence of local minima that hinder convergence.
To further analyze this phenomenon, Fig.~\ref{fig:trajs}~(d) presents the optimization trajectories of iFusion initialized from four distinct azimuth angles (i.e., $0^\circ$, $90^\circ$, $180^\circ$, and $270^\circ$), overlaid on the corresponding 2D MSE loss landscape. The visualization reveals how different starting points lead to different minima. Once a trajectory reaches a local minimum, it typically stagnates. These local minima often arise due to geometric ambiguities in the object. For instance, in Fig.~\ref{fig:trajs}~(a), the object exhibits symmetry between its front and back sides despite textural differences. This geometric characteristic gives rise to two deep valleys in the loss landscape, located 180 degrees apart in azimuth. Finally, Fig.~\ref{fig:trajs}~(c) illustrates the effect of initialization by presenting images generated from Zero123 with poses along the optimization trajectories. This highlights the substantial impact of initial conditions on the quality of the final outcome.

To address the local minima issue caused by object-dependent geometric ambiguities, we introduce a score-based model that guides the optimization toward regions of high data likelihood. This motivates our two-stage optimization framework. In the first stage, a score-based model is trained to learn the score of the data distribution. This provides guidance that steers the optimization away from local minima.
Having escaped poor local minima through the first stage, the second stage employs the pretrained diffusion model with an MSE loss to further refine the pose estimate. By employing these complementary stages, the framework significantly mitigates, or even eliminates, the need for multiple initialization points, which in turn improves the sample efficiency of the gradient-based solver.
Moreover, since the primary goal involves reshaping the loss landscape, we further investigate the energy modeling approach. This approach trains the model to learn an energy function that represents the data distribution. After training, the score is obtained through differentiation of the energy, and optimization is performed via gradient descent. In the experiments, the score-based formulation demonstrates superior convergence behavior and competitive accuracy compared to state-of-the-art (SoTA) methods while requiring fewer samples and reducing inference time. The contributions of this work are summarized as follows:
\vspace{-0.4em}
\setlength{\leftmargini}{10pt}
\begin{itemize}
\item We introduce the \textit{landscape perspective} for analyzing two-view pose estimation. In contrast to prior methods such as energy-based approaches and iFusion, we present the first systematic study examining how optimization landscapes essentially affect the pose estimation process.

\vspace{-0.1em}
\item We propose a score-based method that learns to fundamentally reshape the optimization landscape and gradient field. This approach effectively mitigates local minima without resorting to dense multi-initialization strategies.
\vspace{-0.1em}
\item We conduct comprehensive comparisons with SoTA approaches, achieving performance on par with existing ones while requiring fewer samples and faster inference.
\end{itemize}

\section{Related Work}
\vspace{-0.6em}
\textbf{Pose Estimation.}\hspace{0.5em} Traditional approaches estimate pose by detecting and matching keypoints across images, using either hand-crafted features~\citep{DBLP:journals/pami/LagunaM23, DBLP:conf/eccv/BayTG06, DBLP:conf/bmvc/HarrisS88, DBLP:journals/ijcv/Lowe04, DBLP:conf/eccv/RostenD06} or learned descriptors~\citep{detone2018superpoint, dusmanu2019d2, revaud2019r2d2, DBLP:conf/nips/TyszkiewiczFT20}. The matched correspondences are then used to recover relative poses. While these traditional methods~\citep{DBLP:conf/cvpr/SchonbergerF16} are effective in well-textured and dense-view scenarios, they often struggle under sparse-view or textureless conditions. 
To address these limitations, learning-based methods have emerged that bypass explicit feature matching. Some directly regress camera pose from input images~\citep{DBLP:conf/iccv/KendallGC15,DBLP:conf/cvpr/Sinha0TGL23,tang2024aden, DBLP:conf/cvpr/DongWLCFK025}, while others adopt energy-based formulations~\citep{DBLP:conf/eccv/ZhangRT22,DBLP:conf/3dim/LinZRT24,murphy2021implicit,DBLP:conf/wacv/HoferKMZ23,DBLP:conf/iccvw/HaugaardHI23} or leverage diffusion models~\citep{wang2023posediffusion,hsiao2024confronting}. More recent trends predict dense ray or point maps, as in DUSt3R~\citep{wang2024dust3r}, MASt3R~\citep{mast3r_eccv24}, and VGGT~\citep{DBLP:conf/cvpr/WangCKV0N25}, providing stronger geometric constraints and thus more stable.  Another emerging direction repurposes pretrained pose-conditioned diffusion models in reverse for pose estimation, which we discuss in the following paragraphs.

\noindent\textbf{Exploiting Pose-conditioned Diffusion Models on Pose Estimation.}\hspace{0.5em} Pose-conditioned diffusion models~\citep{DBLP:conf/iccv/LiuWHTZV23, shi2023zero123++, chen2024cascade, sargent2023zeronvs} are generative models fine-tuned from pretrained diffusion models~\citep{DBLP:conf/cvpr/RombachBLEO22} to enable control over camera viewpoints, synthesizing images conditioned on a reference image and a given camera pose. 
Recent works invert these models for pose estimation. Methods such as ID-pose~\citep{cheng2023id} and iFusion~\citep{zhao2024sparse} estimate the relative pose between a reference image and a query image by inverting a pretrained pose-conditioned diffusion model. These approaches iteratively refine the pose by using Zero123 to predict the noise given an image pair and the current pose estimate, then comparing the predicted noise with the actual noise added to the query image. The pose is updated via gradient-based optimization to minimize this noise-space discrepancy.

\begin{figure*}[t]
\vspace{-1.5em}
\centering
\footnotesize
\includegraphics[width=1.0\linewidth]{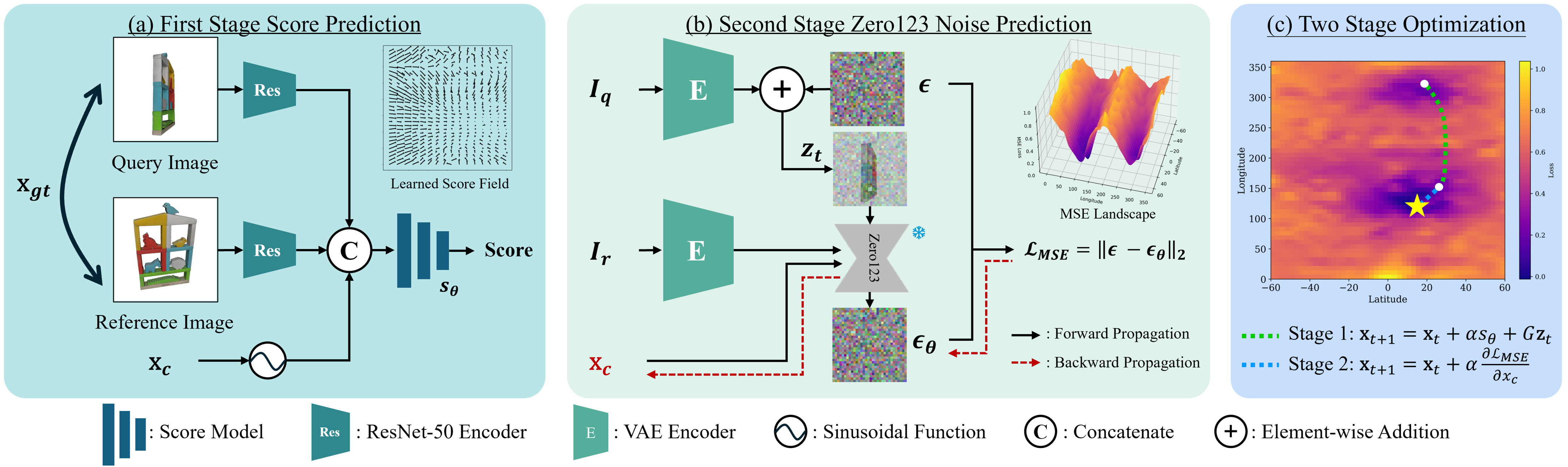}
\vspace{-1.8em}
    \caption{\textbf{Framework Overview.} 
    (a) The first part shows our proposed score network, which uses a ResNet encoder to extract image features. The conditioned pose is encoded via a sinusoidal embedding, and these features are concatenated and fed into a MLP to predict score. The trained score function guides optimization trajectory toward the ground-truth pose, helping avoid local minima in the Zero123 MSE landscape.
    (b) The second stage uses Zero123 to refine via energy-based optimization. Given a pair of reference-query images and a conditioned pose, the frozen Zero123 model estimates the noise. The MSE between predicted and actual noise defines energy, which is minimized via gradient-based optimization to refine the pose. 
    (c) The two stages are combined to form the overall optimization process.
    }
\label{fig:architecture}
\vspace{-1.5em}
\end{figure*}

\vspace{-0.8em}
\section{Preliminary}
\vspace{-0.6em}
\label{sec:preliminary}
This section reviews background concepts relevant to our work, including score-based and energy-based modeling techniques, and framework for leveraging pretrained pose-conditioned diffusion models in camera pose estimation.

To enable score-based learning over empirical data, it is essential to define a differentiable approximation of the underlying data distribution. Let $\{\mathbf{x}^{(i)}\}_{i=1}^N$ denote a set of $i.i.d$ samples, where each $\mathbf{x}^{(i)}\in \mathbb{R}^d$ represents the $i^{\text{th}}$ observation drawn from an unknown distribution $p_{\text{data}}(\mathbf{x})$.
A direct representation of this distribution can be expressed as a mixture of Dirac delta functions:  $\frac{1}{N}\sum_{i=1}^N \delta(\lVert \mathbf{x}-\mathbf{x}^{(i)} \rVert)$, which exactly matches the observed samples but is inherently non-differentiable and thus unsuitable for gradient-based optimization.
To enable a smooth and tractable approximation, Parzen density estimation~\citep{DBLP:journals/neco/Vincent11} smooths each Dirac delta with an isotropic Gaussian kernel: $p_\sigma(\tilde{\mathbf{x}} \mid \mathbf{x}) = \mathcal{N}(\tilde{\mathbf{x}} \mid \mathbf{x},\sigma^2\mathbf{I}_d)$, where $\sigma^2$ controls the degree of smoothing. This formulation yields a differentiable density estimate that forms the foundation for score-based learning.

\noindent\textbf{Score-based Modeling.}\hspace{0.5em} Given the smooth density approximation introduced above, score-based modeling \citep{DBLP:journals/jmlr/Hyvarinen05, DBLP:journals/neco/Vincent11} aims to estimate the score of a data distribution. The score captures the direction that increases the data likelihood and thus provides a principled way to guide optimization toward regions of high probability. The score function is represented by a neural network $s_\theta(\mathbf{x})$ parameterized by $\theta$.  The network is trained using the Denoising Score Matching (DSM) loss \citep{DBLP:journals/neco/Vincent11} $\mathcal{L}_{\mathrm{DSM}}$, expressed as:
\vspace{-0.3em}
\begin{equation}
\scalebox{1.0}{$
    \mathcal{L}_{\mathrm{DSM}}(\theta)=\frac{1}{2}\mathbb{E}_{\tilde{\mathbf{x}}, \mathbf{x}} \left[ 
    \lVert  s_\theta(\tilde{\mathbf{x}}) - \nabla_{\tilde{\mathbf{\mathbf{x}}}} \log p_\sigma (\tilde{\mathbf{x}}\mid \mathbf{x}) \rVert_2^2
    \right],
$}
\label{eq:dsm_loss}
\vspace{-0.3em}
\end{equation}
where $ \nabla_{\tilde{\mathbf{x}}} \log p_{\sigma}(\tilde{\mathbf{x}} \mid \mathbf{x}) $ is the denoising direction that can be computed analytically.
Minimizing Eq.~(\ref{eq:dsm_loss}) trains $s_{\theta}$ to approximate the true score $\nabla_{\tilde{\mathbf{x}}}\log p_\sigma(\tilde{\mathbf{x}})$ of the smoothed data distribution~\citep{DBLP:journals/neco/Vincent11}. Once trained, the score can be utilized in Langevin dynamics to iteratively sample from this distribution through: $ \tilde{\mathbf{x}}_t=\tilde{\mathbf{x}}_{t-1} + \frac{\alpha}{2}s_\theta(\tilde {\mathbf{x}}_{t-1}) + \sqrt{\alpha} \mathbf{z}_t $, where $ \mathbf{z}_t \sim \mathcal{N}(0, I) $ represents Gaussian noise, $ \alpha $ is the step size, and $t$ is iteration number. In our context, this learned score function later serves as a guide for optimization, steering updates toward more probable poses.

\noindent\textbf{Energy-Based Modeling.}\hspace{0.5em} Energy-based models (EBMs) provide an alternative but closely related perspective. They represent data distributions using an energy function $\mathcal{E}(\mathbf{x})$, where low-energy regions correspond to high-likelihood samples: $p(\mathbf{x})=\frac{1}{Z} \exp(-\mathcal{E}(\mathbf{x}))$, where $Z$ denotes the normalization constant. The score is the gradient of the log-density, which can be expressed as $s(\mathbf{x})=\nabla_{\mathbf{x}}\log p(\mathbf{x})=-\nabla_{\mathbf{x}}\mathcal{E}(\mathbf{x})$. This relationship shows that EBMs and score-based models define equivalent gradient fields over the data manifold, while EBMs model the energy function itself, score-based models directly approximate its gradient.

\noindent\textbf{Inverting Pose-conditioned Diffusion Model.}\hspace{0.5em} Zero123 generates novel views conditioned on a reference image and a relative camera pose. Let $I_r$ be the reference image and $I_q$ the query image. The query image is encoded by a VAE encoder: $\mathbf{z}=\mathrm{E}(I_q)$, and Gaussian noise $\epsilon\sim\mathcal{N}(0,I)$ is added to produce the noisy latent $\mathbf{z}_t$. Zero123 is trained to predict this noise:
$\mathcal{L}(I_q,(I_r,T))=\mathbb{E}_{\mathbf{z},\epsilon,t}\left[ \lVert \epsilon-\epsilon_\theta(\mathbf{z}_t,t,(I_r, T)) \rVert_2^2 \right]$, where the condition includes the reference image $I_r$ and the relative camera pose $T$. 
Leveraging this formulation, methods such as ID-pose and iFusion invert Zero123 by treating the camera pose $T$ as an optimization variable while freezing the pretrained diffusion parameter $\theta$. The goal is to estimate the relative pose $\hat T_{r\to q}$ that minimize the diffusion denoising MSE loss:
\vspace{-0.5em}
\begin{equation}
\label{eq:ifusion_optimization_formula}
\scalebox{0.95}{$\hat T_{r\to q}=\underset{T\in SE(3)}{\text{argmin}}\mathcal{L}(I_q,(I_r,T))+\mathcal{L}(I_r,(I_q,T^{-1}))$},
\vspace{-0.5em}
\end{equation}
and gradients of the loss with respect to $T$ are used to iteratively update the pose via gradient-based optimization.


\vspace{-0.5em}
\section{Methodology}
\label{sec:methodology}
\subsection{Problem Formulation}
\vspace{-0.5em}
Let the dataset consists of $N$ paired samples $\{(\mathbf{x}^{(i)}, \mathbf{y}^{(i)})\}_{i=1}^N$, where each pair includes a relative camera pose $\mathbf{x}^{(i)}$ and an image pair $\mathbf{y}^{(i)}=(I^{(i)}_r, I^{(i)}_q)$. Here, $I^{(i)}_r$ is the reference image and $I^{(i)}_q$ is the query image.
We parameterize the camera pose in spherical coordinates $(\Theta, \Phi, \rho)$, following prior works such as iFusion and Zero123. The relative pose between the two images is defined as the difference between query and reference poses: $\mathbf{x} = (\Theta_{q} -\Theta_{r}, \Phi_{q} - \Phi_{r}, \rho_{q} - \rho_{r})$. The objective is to estimate the transformation $\mathbf{x}$ from the provided image pair $\mathbf{y}$. It is assumed that each image pair $\mathbf{y}^{(i)}$ corresponds to a unique ground-truth relative pose $\mathbf{x}^{(i)}$. Formally, this implies that the conditional data distribution is deterministic and can be expressed as a Dirac delta function: $p(\mathbf{x}\mid \mathbf{y}^{(i)})=\delta (\mathbf{x}-\mathbf{x}^{(i)})$.

\begin{figure*}[t]
\centering
\footnotesize
\includegraphics[width=1.0\linewidth]{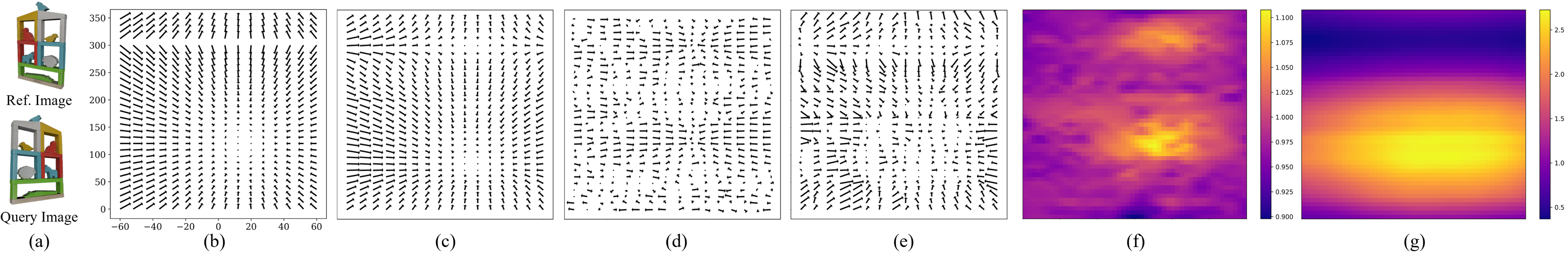}
\vspace{-1.em}
    \caption{
    \textbf{Toy Example.} 
    (a) Reference and query images; (b) Oracle score field; (c) Score field from our score-based model; (d) Score field from Zero123 MSE; 
    (e) Score field from our energy-based model; (f) Probability landscape from Zero123 MSE loss; (g) Probability landscape from our energy-based model. 
    The landscapes in (f) and (g) represent the probability distribution and are plotted as $\exp(-\mathcal{E}_\theta(\mathbf{x}))$.
    }
\label{fig:toy_examples}
\vspace{-1.em}
\end{figure*}

\subsection{The Proposed Framework}
\label{sec:proposed_framework}
\vspace{-0.5em}
In this section, we introduce a two-stage optimization framework for estimating camera poses while mitigating the effect of local minima in the Zero123 energy landscape. 

\noindent\textbf{Framework Overview.}\hspace{0.5em} The proposed framework, as illustrated in Fig.~\ref{fig:architecture}, comprises two stages: \textit{the score-based optimization stage} and \textit{the energy-based refinement stage}.
In the first stage, our framework employs a score network $s_\theta (I_r,I_q,\tilde{\mathbf{x}})$ that predicts the update direction of the pose for a reference image $I_r$, a query image $I_q$, and a noised pose $\tilde{\mathbf{x}}$. The underlying principle is that the score model learns to approximate the gradient of the log-probability density of plausible poses conditioned on the image pair. Through iterative updates along this learned gradient, the pose is encouraged to move toward high-probability regions of the pose space, which effectively escapes local minima. More specifically, the score network adopts a lightweight design: image features are extracted by a ResNet-50~\citep{DBLP:conf/cvpr/HeZRS16} backbone, while the conditional pose is encoded via sinusoidal embeddings. The concatenated features are processed by a three-layer MLP to predict the score function.


Once the pose has been guided toward a geometrically consistent region, the second stage employs the pretrained Zero123 model to refine it through energy-based optimization. In particular, Gaussian noise is injected into the latent representation of the query image to obtain a noisy latent $\mathbf{z}_t$. The Zero123 model then estimates the noise conditioned on the reference image and the current pose estimate. The MSE between the predicted and original noise serves as the energy function, where the gradient of this energy with respect to the pose provides a refinement direction. This energy-driven gradient descent further aligns the pose with the cross-view consistency encoded in the Zero123 model.

This two-stage pipeline forms a coherent optimization strategy: the score model first provides global guidance to avoid suboptimal local minima, while the diffusion-based energy model subsequently delivers fine-grained local corrections. Both stages rely on gradient-based updates but differ in the manner through which the gradient is obtained. The first stage derives it explicitly from the learned score, while the second obtains it implicitly from the energy loss.


\noindent\textbf{Training Objective.}\hspace{0.5em} 
Our training objective for the score model $s_\theta$ follows the denoising score matching (DSM) principle, extended to the conditional setting:
\vspace{-0.6em}
\begin{equation}
\scalebox{0.93}{$\mathcal{L}(\theta) = \frac{1}{2} \mathbb{E}_{\mathbf{x}, \mathbf{y}} \mathbb{E}_{\tilde{\mathbf{x}}\sim \mathrm{U}}  \left\| s_\theta(\tilde{\mathbf{x}}, \mathbf{y}) - \nabla_{\tilde{\mathbf{x}}} \log p_\sigma(\tilde{\mathbf{x}} \mid \mathbf{x}, \mathbf{y}) \right\|_2^2.
$}
\vspace{-0.6em}
\label{eq:train_obj_score}
\end{equation}
Since our score model operates in the low-dimensional pose space, we adopt a simplified objective compared to NCSN~\citep{DBLP:conf/nips/SongE19}. Specifically, $\tilde{\mathbf{x}}$ is sampled from a uniform distribution $\mathrm{U}$, and the noise scale is fixed at $\sigma = 1$, allowing the model to disregard noise-level conditioning. This uniform sampling enables the model to learn a score function that captures the global gradient structure over the entire pose space, rather than focusing only on a local neighborhood around $\mathbf{x}$.
Despite this simplification, we show in Section~\ref{sec:theoretical_justification} that the optimal solution remains theoretically equivalent to that obtained with a Gaussian kernel.
Implementation details and hyperparameters are provided in the Appendix~\ref{sec:implementation_details}. 
We further compare our score-based formulation with an energy-based alternative in Appendix~\ref{sec:derivation_details}.


\begin{figure*}[t]
\centering
\footnotesize
\captionsetup{skip=1pt, , belowskip=1pt}
\includegraphics[width=1.0\linewidth]{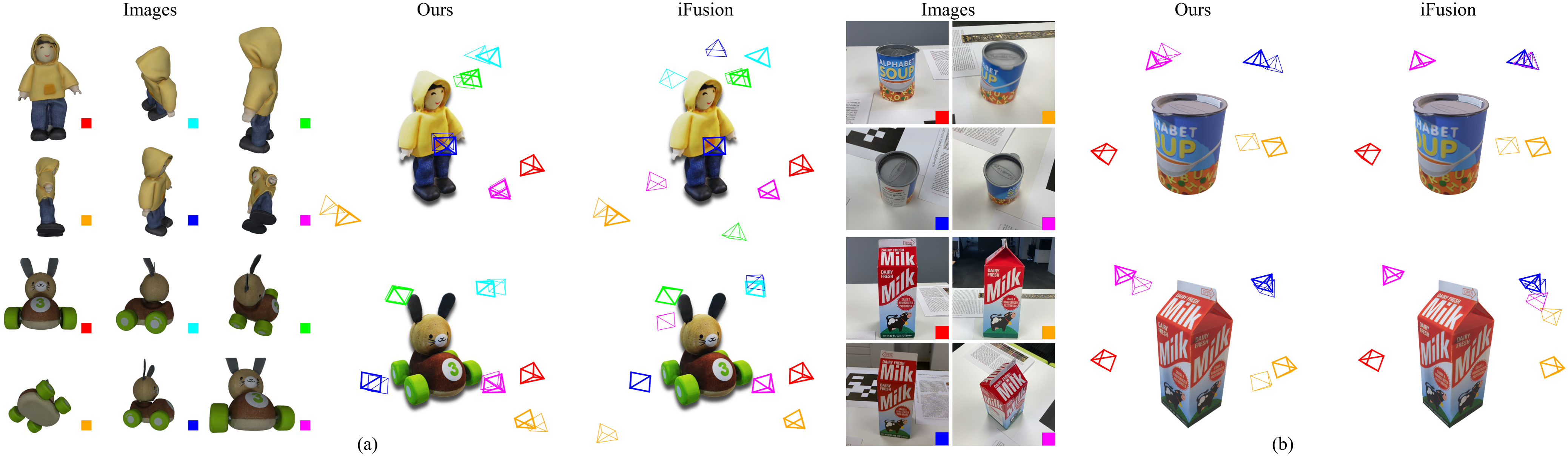}
    \caption{\textbf{Qualitative Results.} Visualization of predicted camera poses (thin) compared to ground truth poses (bold). For each object, we randomly select two initial viewpoints and estimate the relative poses of all target views from a reference image, shown in {\color{red}red}. (a) Results on GSO objects: our method consistently converges to the correct pose, while iFusion often gets stuck in local minima, leading to incorrect predictions. (b) Results on HOPEv2 real-world objects: our method accurately recovers the correct poses despite strong geometric symmetries, as the distinct textures allow effective disambiguation, whereas iFusion frequently converges to incorrect local optima.}
\label{fig:qualitative_result}
\vspace{-1.0em}
\end{figure*}

\subsection{Two Stage Optimization Process}
\vspace{-0.5em}
After training, the learned score model enables guidance of arbitrary initial poses toward higher-density regions of the pose distribution. Once the optimization escapes poor local minima, the Zero123 MSE loss further refines the estimated pose. Since precisely determining when a local minima has been escaped is challenging in practice, we adopt a fixed iteration threshold, after which the optimization proceeds using the MSE gradient. Consequently, the overall process consists of two distinct stages. In the first stage, the pose is updated using the learned score model to guide the initial optimization toward regions of higher data likelihood:
\vspace{-0.5em}
\begin{equation}
    \tilde{\mathbf{x}}_{t}=\tilde{\mathbf{x}}_{t-1}+\alpha s_\theta(\tilde{\mathbf{x}}_{t-1}, \mathbf{y}) + G \mathbf{z}_t, \quad \mathbf{z}_t\sim \mathcal{N}(0,\mathbf{I}_3),
    \label{eq:update_function}
\vspace{-0.5em}
\end{equation}
\noindent where $G=\text{diag}(\gamma_1,\gamma_2,\gamma_3)$ controls the noise scale for each coordinate. This update resembles Langevin dynamics, where the learned score provides a drift toward high-likelihood regions and the Gaussian noise encourages exploration of the pose space. This dynamics ensures that the norm of the expected pose error decays exponentially.
\vspace{-0.8em}
\begin{equation}
\lVert\mathbb{E}\!\left[ \tilde{\mathbf{x}}_t - \mathbf{x}_{\mathrm{gt}} \right]\rVert = M (1 - \alpha)^t, 
\quad 
\mathrm{Var}\!\left[\tilde{\mathbf{x}}_{t}\right] \approx \frac{G^2}{2\alpha},
\label{eq:decay_converge}
\vspace{-0.5em}
\end{equation}
\noindent where $M$ is a constant depending on the initial distance. A detailed proof is provided in the Appendix~\ref{sec:derivation_details}.
In the second stage, the pretrained Zero123 model is employed as an energy function, and gradient-based methods is performed to solve the optimization problem defined in Eq.~(\ref{eq:ifusion_optimization_formula}).

\noindent \textbf{Joint Reasoning across Multiple Views.}\hspace{0.5 em} 
A straightforward extension of a two-view method to the multi-view setting is to process each image pair independently; however, this ignores multi-view consistency.
To address this, we formulate a unified objective that performs energy-based optimization in a high-dimensional pose space, as shown in Eq.~(\ref{eq:energy_optim_multiview}). Enforcing global consistency allows reliable relations to correct erroneous ones, improving robustness.
\vspace{-0.5em}
\begin{equation}
\label{eq:energy_optim_multiview}
\scalebox{0.88}{$\mathcal{\hat{T}}
= \underset{\{T_1, \dots, T_n\} \subset SE(3)}{\arg\min}\ 
\sum_{i=1}^N\sum_{j \ne i} 
\mathcal{L}\!\left(I^{(j)},\,
(I^{(i)},\,
T_i^{-1} T_j) \right),$}
\vspace{-0.5em}
\end{equation}
where $\{T_1, \dots,T_n\}$ denotes the set of absolute camera poses for all views.
Despite its advantages, optimizing Eq.~(\ref{eq:energy_optim_multiview}) is challenging. The solution space grows exponentially with the number of views, making multi-start strategies computationally prohibitive and prone to local minima.

To address this, we extend our two-stage framework to the multi-view scenario. Given $N$ images, the first stage uses the learned score function to infer pairwise relative poses $\mathcal{T}=\{T_{i \to j}\}_{i\ne j}$. We then perform a global optimization to obtain a consistent set of absolute poses $\overline{\mathcal{T}}=\{\overline{T}_{i}\}_{i=1}^N$, where $\overline{T}_{i\to j}=\overline{T}_i^{-1}\overline{T}_j$. This reparameterization removes redundancy and enforces global consistency across all views.
The resulting estimate $\overline{\mathcal{T}}$ provides a strong initialization for the subsequent refinement. Then we apply a Zero123-based energy optimization, using Eq.~(\ref{eq:energy_optim_multiview}), to refine the poses and obtain the final transformation set $\mathcal{\hat{T}}$.

\subsection{Theoretical Justification}
\label{sec:theoretical_justification}
\vspace{-0.5em}
We provide a theoretical analysis of the objective in Eq.~(\ref{eq:train_obj_score}).

\noindent\textbf{Proposition 1.} 
Consider the objective
\vspace{-0.2em}
\begin{equation}
    \scalebox{1.0}{$\mathcal{L}(\theta) = \frac{1}{2}  \mathbb{E}_{\mathbf{y}, \mathbf{x}, \tilde{\mathbf{x}}}  \left\| s_\theta(\tilde{\mathbf{x}}, \mathbf{y}) - \nabla_{\tilde{\mathbf{x}}} \log p_\sigma(\tilde{\mathbf{x}} \mid \mathbf{x}) \right\|_2^2$,}
    \label{eq:DSM_obj}
\vspace{-0.2em}
\end{equation}
\noindent where the expectation is taken over $p(\mathbf{y}), p(\mathbf{x}\mid \mathbf{y})$, and $p(\tilde{\mathbf{x}}\mid \mathbf{x})$.
If $p_\sigma(\tilde{\mathbf{x}}\mid \mathbf{x})=\mathcal{N}(\tilde{\mathbf{x}}; \mathbf{x},\sigma^2 I)$, then the optimal solution $s^\star(\tilde{\mathbf{x}},\mathbf{y})$ for fixed $(\tilde{\mathbf{x}},\mathbf{y})$ is
\vspace{-0.2em}
\begin{equation}
    s^\star(\tilde{\mathbf{x}},\mathbf{y})=\mathbb{E}_{\mathbf{x}\sim p(\mathbf{x}\mid \mathbf{y},\tilde{\mathbf{x}})}[\nabla_{\tilde{\mathbf{x}}}\log p_\sigma(\tilde{\mathbf{x}}\mid \mathbf{x})].
\vspace{-0.2em}
\end{equation}
\noindent Equivalently, in integral form,
\vspace{-0.2em}
\begin{equation}
    s^\star(\tilde{\mathbf{x}},\mathbf{y})
    = \frac{\displaystyle\int (\mathbf{x}-\tilde{\mathbf{x}} )\,p(\tilde{\mathbf{x}}\mid \mathbf{x})\,p(\mathbf{x}\mid \mathbf{y})\,d\mathbf{x}}
    {\sigma^2 \displaystyle\int p(\tilde{\mathbf{x}}\mid \mathbf{x})\,p(\mathbf{x}\mid \mathbf{y})\,d\mathbf{x}}.
\vspace{-0.2em}
\end{equation}

To simplify the sampling strategy, consider the case where $\tilde{\mathbf{x}}$ is sampled from a uniform distribution $\mathrm{U}$. This motivates the following lemma.

\noindent\textbf{Lemma 1.} 
Given $\mathbf{y}$ and $\tilde{\mathbf{x}}\sim \mathrm{U}$, where $\mathrm{U}$ denotes a uniform distribution independent of $\mathbf{x}$, the optimal solution of the loss function $\mathcal{L}(\theta)$ in Eq.~(\ref{eq:DSM_obj}) is
\vspace{-0.3em}
\begin{align}
    s_U^\star(\tilde{\mathbf{x}},\mathbf{y})&=\mathbb{E}_{\mathbf{x}\sim p(\mathbf{x}\mid \mathbf{y})}[\nabla_{\tilde{\mathbf{x}}}\log p_\sigma(\tilde{\mathbf{x}}\mid \mathbf{x})] \notag \\
    &=\frac{1}{\sigma^2}\int (\mathbf{x}-\tilde{\mathbf{x}})p(\mathbf{x}\mid \mathbf{y}) \, d\mathbf{x}.
\vspace{-0.3em}
\end{align}
\noindent\textbf{Lemma 2.} 
In general, $s^\star(\tilde{\mathbf{x}},\mathbf{y})\ne s_\mathrm{U}^\star(\tilde{\mathbf{x}},\mathbf{y})$, except in the special case where $p(\mathbf{x}\mid \mathbf{y})$ collapses to a Dirac delta distribution, i.e., $p(\mathbf{x}\mid \mathbf{y}^{(i)})=\delta(\mathbf{x}-\mathbf{x}^{(i)})$.

Lemma~2 shows that, under the assumption of a unique solution for each conditional image pair, the simplified objective shares the same optimum as Eq.~(\ref{eq:DSM_obj}). Details proofs of Proposition~1 as well as Lemma~1 and 2 are provided in Appendix~\ref{sec:derivation_details} for completeness.

\begin{table*}[t]
\vspace{-1em}
\centering
\caption{\textbf{Evaluation results on the synthetic dataset.} Results on the GSO and OO3D datasets show that our two-stage optimization framework improves success rate and recall across thresholds. \textbf{\textcolor{red}{Red}} indicates our best result, and \textbf{\textcolor{blue}{blue}} denotes the second best result.}
\vspace{-1em}
\renewcommand{\arraystretch}{1.2} 
    \begin{adjustbox}{width=1.0\textwidth} 
    \begin{tabular}{l l c c c c c c c c c c c c c c}
    \specialrule{1.5pt}{0pt}{0pt}
    \multirow{2}{*}{\textbf{Dataset}} & \multirow{2}{*}{\textbf{Method}} & \multicolumn{2}{c}{\textbf{@5}} & \multicolumn{2}{c}{\textbf{@15}} & \multicolumn{2}{c}{\textbf{@30}} & \multicolumn{2}{c}{\textbf{@5}} & \multicolumn{2}{c}{\textbf{@15}} & \multicolumn{2}{c}{\textbf{@30}} & \multirow{2}{*}{\textbf{Rot. $\downarrow$}} & \multirow{2}{*}{\textbf{Trans. $\downarrow$}} \\
    \cmidrule(lr){3-4} \cmidrule(lr){5-6} \cmidrule(lr){7-8} \cmidrule(lr){9-10} \cmidrule(lr){11-12} \cmidrule(lr){13-14}
    & & \textbf{R $\uparrow$} & \textbf{R(R) $\uparrow$} & \textbf{R $\uparrow$} & \textbf{R(R) $\uparrow$} & \textbf{R $\uparrow$} & \textbf{R(R) $\uparrow$} & \textbf{SR $\uparrow$} & \textbf{SR(R) $\uparrow$} & \textbf{SR $\uparrow$} & \textbf{SR(R) $\uparrow$} & \textbf{SR $\uparrow$} & \textbf{SR(R) $\uparrow$} & & \\

    \specialrule{1.2pt}{0pt}{0pt}
    \multirow{5}{*}{\textbf{GSO}} & \textbf{DUSt3R}~\citep{mast3r_eccv24} & 0.530 & 0.534 & 0.903 & \textbf{\textcolor{blue}{0.923}} & \textbf{\textcolor{red}{0.957}} & \textbf{\textcolor{red}{0.986}}  & - & - & -& - & - & - & 4.63 & 0.053 \\
    & \textbf{VGGT}~\citep{DBLP:conf/cvpr/WangCKV0N25} & \textbf{\textcolor{red}{0.752}} & \textbf{\textcolor{red}{0.800}} & 0.866 & \textbf{\textcolor{red}{0.945}} & 0.869 &  \textbf{\textcolor{blue}{0.960}} & - & - & -& - & - & - & \textbf{\textcolor{red}{2.14}} & \textbf{\textcolor{blue}{0.050}} \\
    & \textbf{ID-Pose}~\citep{cheng2023id} & 0.223 & 0.247 & 0.541 & 0.624 & 0.607 &  0.723 & 0.039 & 0.049 & 0.118 & 0.152 & 0.146 & 0.201 & 10.29 & 0.134 \\
    & \textbf{iFusion}~\citep{wu2023ifusion} & \textbf{\textcolor{blue}{0.700}} & \textbf{\textcolor{blue}{0.704}} & \textbf{\textcolor{blue}{0.904}}  & {0.916}  & {0.918} & {0.938} & \textbf{\textcolor{blue}{0.275}}  & \textbf{\textcolor{blue}{0.278}} & \textbf{\textcolor{blue}{0.365}} & \textbf{\textcolor{blue}{0.374}} & \textbf{\textcolor{blue}{0.382}} & \textbf{\textcolor{blue}{0.398}} & \textbf{\textcolor{blue}{3.07}} & \textbf{\textcolor{red}{0.035}} \\
    & \textbf{Ours} & {0.645} & {0.650} & \textbf{\textcolor{red}{0.907}} & {0.921} & \textbf{\textcolor{blue}{0.927}} & {0.945} &  \textbf{\textcolor{red}{0.585}}& \textbf{\textcolor{red}{0.591}} & \textbf{\textcolor{red}{0.811}} & \textbf{\textcolor{red}{0.840}} & \textbf{\textcolor{red}{0.836}} & \textbf{\textcolor{red}{0.878}} & {3.63} & {0.051} \\
    \hline
    \multirow{5}{*}{\textbf{OO3D}} & \textbf{DUSt3R}  & 0.395 & 0.404 & 0.791 & 0.832 & \textbf{\textcolor{blue}{0.889}} & \textbf{\textcolor{blue}{0.969}}  & - & - & -& - & - & - & 6.49 & \textbf{\textcolor{blue}{0.054}} \\
    & \textbf{VGGT} & 0.477 & \textbf{\textcolor{blue}{0.514}} & 0.739 & 0.830 & 0.757 &  0.875 & - & - & -& - & - & - & \textbf{\textcolor{blue}{4.81}} & {0.085} \\
    & \textbf{ID-Pose}~\citep{cheng2023id} & 0.134 & 0.146 & 0.454 & 0.531 & 0.546 & 0.694 & 0.022 & 0.028 & 0.084 & 0.111 & 0.113 & 0.167 & 13.79 & 0.147 \\
    &  \textbf{iFusion} & \textbf{\textcolor{red}{0.516}} & \textbf{\textcolor{red}{0.523}} & \textbf{\textcolor{red}{0.841}} & \textbf{\textcolor{blue}{0.866}} & 0.882 & 0.930 & \textbf{\textcolor{blue}{0.189}} & \textbf{\textcolor{blue}{0.192}} & \textbf{\textcolor{blue}{0.306}} & \textbf{\textcolor{blue}{0.316}} & \textbf{\textcolor{blue}{0.332}} & \textbf{\textcolor{blue}{0.359}} & \textbf{\textcolor{red}{4.76}} & \textbf{\textcolor{red}{0.047}} \\
    & \textbf{Ours} & \textbf{\textcolor{blue}{0.479}} & {0.489} & \textbf{\textcolor{blue}{0.838}} & \textbf{\textcolor{red}{0.875}} & \textbf{\textcolor{red}{0.905}} & \textbf{\textcolor{red}{0.970}} & \textbf{\textcolor{red}{0.447}} & \textbf{\textcolor{red}{0.464}} & \textbf{\textcolor{red}{0.780}} & \textbf{\textcolor{red}{0.842}} & \textbf{\textcolor{red}{0.848}} & \textbf{\textcolor{red}{0.949}} & {5.15} & 0.055 \\
    \specialrule{1.5pt}{0pt}{0pt}
    \end{tabular}
    \end{adjustbox}
\vspace{-0.5em}
\label{tab:eval_results_myGSO_myOO3D}
\end{table*}

\subsection{Toy Example}
We present a toy example trained on a single object from the GSO~\citep{DBLP:conf/icra/DownsFKKHRMV22} dataset. This example demonstrates that both score-based and energy-based modeling can learn the underlying data distribution. As illustrated in Fig.~\ref{fig:toy_examples}, we visualize the score field for the score-based method and the probability density $\exp(-\mathcal{E}_\theta(\mathbf{x}))$ for the energy-based method. For the latter, we also display the induced score field obtained via gradient computation. Fig.~\ref{fig:toy_examples}~(f) depicts the probability density computed from the Zero123 MSE landscape, which exhibits two modes due to the object's geometry, and Fig.~\ref{fig:toy_examples}~(d) visualizes the corresponding score field. In contrast, Fig.~\ref{fig:toy_examples}~(c) presents the score field from the score-based modeling, which aligns well with the oracle score field in Fig.~\ref{fig:toy_examples}~(b). Fig.~\ref{fig:toy_examples}~(g) reveals the probability density learned by the energy-based model and demonstrates a smooth and coherent landscape. However, its corresponding score field in Fig.~\ref{fig:toy_examples}~(e) appears noisy and inferior to that of the score-based method. We conjecture that this limitation arises from the indirect prediction process, which requires gradient computation on the energy function. A more comprehensive comparison of these two modeling approaches is provided in Section~\ref{sec:ablation_study}.


\section{Experimental Results}
\label{sec:experimental_results}
\vspace{-0.5em}
\subsection{Experimental Setups}
\label{sec:experimental_setups}
\vspace{-0.5em}

\textbf{Dataset.}\hspace{1em} 
Following the setup in iFusion~\citep{wu2023ifusion}, we adopt the GoogleScannedObject (GSO)~\citep{DBLP:conf/icra/DownsFKKHRMV22} and OmniObject3D (OO3D)~\citep{DBLP:conf/cvpr/WuZFWRPWYWQLL23} 3D model datasets for our experiments. 
Note that additional dataset details are provided in Appendix~\ref{sec:implementation_details}. For real-world evaluation, we use the HOPEv2~\citep{DBLP:conf/iros/TyreeTTCMSB22} dataset from the BOP Challenge~\citep{DBLP:journals/corr/abs-2504-02812}, which contains 28 grocery objects captured in 50 scenes across diverse household and office environments under varying lighting conditions.


\noindent\textbf{Evaluation Metrics.}\hspace{1em} We evaluate our method using the following metrics: Recall (R), Success Rate (SR), Rotation Error (Rot.), and Translation Error (Trans.). Recall measures the proportion of final predictions that satisfy predefined thresholds. Specifically, in gradient-based optimizer, we typically sample $N$ initial poses and select the lowest loss among the $N$ trials. Recall considers this best prediction. In contrast, Success Rate evaluates all $N$ predictions and reports the percentage that meet the thresholds, thereby reflecting the method's robustness to varying initializations. For both Success Rate and Recall, we adopt rotation thresholds of $5^\circ$, $15^\circ$, $30^\circ$, and a translation distance threshold of $0.2$. To isolate rotational performance, we also report Rotation Recall and Rotation Success Rate, denoted as R(R) and SR(R), which assess accuracy solely based on rotation thresholds. Finally, we compute Rotation Error and Translation Error as the median across all evaluated samples to reduce sensitivity to outliers caused by convergence failures.

\noindent\textbf{Evaluations.}\hspace{1em} 
We evaluate our framework on camera pose estimation tasks and compare it against ID-Pose~\citep{cheng2023id}, iFusion~\citep{wu2023ifusion}, DUSt3R~\citep{wang2024dust3r}, and VGGT~\citep{DBLP:conf/cvpr/WangCKV0N25}. We consider two initialization strategies. The first strategy employs nine uniformly distributed poses with latitudes of $-30^\circ$, $0^\circ$, and $30^\circ$, and longitudes from $0^\circ$ to $360^\circ$ in $120^\circ$ increments. The second strategy varies the number of randomly sampled initial poses to evaluate the robustness through recall.

\begin{table*}[t]
\centering
\caption{\textbf{Evaluation results on the HOPEv2 dataset.}\hspace{1em}
Our method maintains strong performance under real-world conditions, as reflected by recall and success rate. \textbf{\textcolor{red}{Red}} indicates the best result, and \textbf{\textcolor{blue}{blue}} the second best.
}
\vspace{-1em}

\renewcommand{\arraystretch}{1.2}
\begin{adjustbox}{width=\textwidth}
\begin{tabular}{l l c c c c c c c c c c c c c c}
\specialrule{1.5pt}{0pt}{0pt}
\multirow{2}{*}{\textbf{Dataset}} & \multirow{2}{*}{\textbf{Method}} & \multicolumn{2}{c}{\textbf{@5}} & \multicolumn{2}{c}{\textbf{@15}} & \multicolumn{2}{c}{\textbf{@30}} & \multicolumn{2}{c}{\textbf{@5}} & \multicolumn{2}{c}{\textbf{@15}} & \multicolumn{2}{c}{\textbf{@30}} & \multirow{2}{*}{\textbf{Rot. $\downarrow$}} & \multirow{2}{*}{\textbf{Trans. $\downarrow$}} \\
\cmidrule(lr){3-4} \cmidrule(lr){5-6} \cmidrule(lr){7-8} \cmidrule(lr){9-10} \cmidrule(lr){11-12} \cmidrule(lr){13-14}
& & \textbf{R $\uparrow$} & \textbf{R(R) $\uparrow$} & \textbf{R $\uparrow$} & \textbf{R(R) $\uparrow$} & \textbf{R $\uparrow$} & \textbf{R(R) $\uparrow$} & \textbf{SR $\uparrow$} & \textbf{SR(R) $\uparrow$} & \textbf{SR $\uparrow$} & \textbf{SR(R) $\uparrow$} & \textbf{SR $\uparrow$} & \textbf{SR(R) $\uparrow$} & & \\
\specialrule{1.2pt}{0pt}{0pt}
\multirow{3}{*}{\textbf{HOPEv2}~\citep{DBLP:conf/iros/TyreeTTCMSB22}}& \textbf{VGGT} & \textbf{\textcolor{blue}{0.208}} & \textbf{\textcolor{red}{0.279}} & \textbf{\textcolor{blue}{0.571}} & \textbf{\textcolor{red}{0.798}} & \textbf{\textcolor{blue}{0.631}} & \textbf{\textcolor{red}{0.893}} & - & - & - & - & - & - & \textbf{\textcolor{red}{8.10}} & \textbf{\textcolor{blue}{0.132}} \\
& \textbf{iFusion} & 0.095 & 0.107 & 0.411 & 0.506 & 0.494 & 0.619 & \textbf{\textcolor{blue}{0.040}} & \textbf{\textcolor{blue}{0.048}} & \textbf{\textcolor{blue}{0.169}} & \textbf{\textcolor{blue}{0.222}} & \textbf{\textcolor{blue}{0.206}} & \textbf{\textcolor{blue}{0.291}} & 14.78 & 0.151 \\
& \textbf{Ours} & \textbf{\textcolor{red}{0.214}} & \textbf{\textcolor{blue}{0.214}} & \textbf{\textcolor{red}{0.679}} & \textbf{\textcolor{blue}{0.702}} & \textbf{\textcolor{red}{0.851}} & \textbf{\textcolor{blue}{0.887}} & \textbf{\textcolor{red}{0.164}} & \textbf{\textcolor{red}{0.165}} & \textbf{\textcolor{red}{0.534}} & \textbf{\textcolor{red}{0.546}} & \textbf{\textcolor{red}{0.786}} & \textbf{\textcolor{red}{0.837}} & \textbf{\textcolor{blue}{8.96}} & \textbf{\textcolor{red}{0.059}} \\
\specialrule{1.5pt}{0pt}{0pt}
\end{tabular}
\end{adjustbox}
\vspace{-0.5em}
\label{tab:eval_results_HOPEv2}
\end{table*}


\vspace{-0.em}
\begin{figure}
    \centering
    \includegraphics[width=\linewidth]{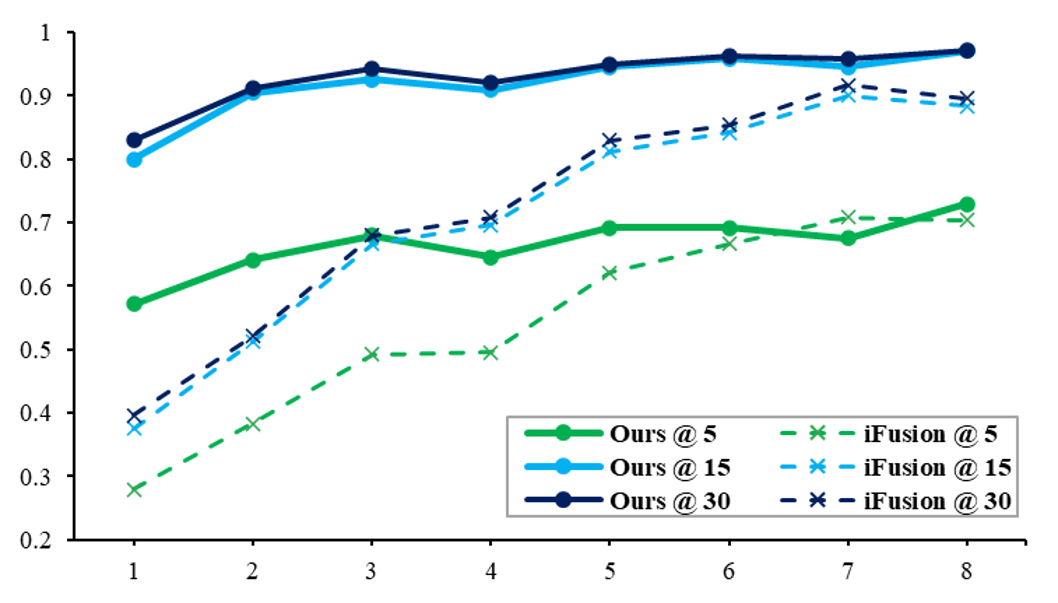}
    \vspace{-1em}
    \caption{\textbf{Evaluation under varying numbers of samples.}
    The figure compares our framework and iFusion in recall for various numbers of random initial poses. The x-axis represents the number of random initial poses, and the y-axis represents the recall.}
    \label{fig:num_samples_recall}
    \vspace{-1em}
\end{figure}

\vspace{-0.5em}
\subsection{Quantitative Results}
\vspace{-0.5em}

Table~\ref{tab:eval_results_myGSO_myOO3D} presents the evaluation results on the GSO and OO3D datasets. Our method achieves substantial improvements in success rate, which demonstrates that the proposed score-based framework effectively mitigates the sample inefficiency inherent in directly applied gradient-based optimization with the Zero123 MSE loss. This improvement is particularly significant as it addresses a fundamental limitation of existing diffusion-based pose estimation methods. Moreover, our method achieves comparable performance to other SoTA methods across recall, rotation error, and translation error metrics. Table~\ref{tab:eval_results_HOPEv2} further reports the evaluation results on real-world data from the HOPEv2 dataset. The results demonstrate robust performance under various challenging real-world conditions, including lighting variation, diverse object textures, material effects, and natural image blur. 
Furthermore, Table~\ref{tab:eval_results_unseen} presents the performance of our framework on unseen objects. The second-stage refinement leverages the strong generative capability of Zero123 to effectively guide the optimization trajectory toward the correct pose. As a result, our approach achieves performance on par with SoTA methods, even though the first-stage score model is trained on a more limited dataset compared to VGGT. This outcome highlights the effectiveness of our two-stage framework in leveraging pretrained generative priors for robust pose estimation on novel objects.


\begin{table}[t]
\centering
\caption{\textbf{Evaluation on unseen objects}. We sample 10 additional objects from the GSO dataset that were not included in training.}
\vspace{-1.0em}
\renewcommand{\arraystretch}{1.2}
\begin{adjustbox}{width=\linewidth}
\begin{tabular}{l c c c c c c c c}
\specialrule{1.5pt}{0pt}{0pt}
\multirow{2}{*}{\textbf{Method}} & \multicolumn{2}{c}{\textbf{@5}} & \multicolumn{2}{c}{\textbf{@15}} & \multicolumn{2}{c}{\textbf{@30}} & \multirow{2}{*}{\textbf{Rot. $\downarrow$}} & \multirow{2}{*}{\textbf{Trans. $\downarrow$}} \\
\cmidrule(lr){2-3} \cmidrule(lr){4-5} \cmidrule(lr){6-7}
& \textbf{R $\uparrow$} & \textbf{R(R) $\uparrow$} & \textbf{R $\uparrow$} & \textbf{R(R) $\uparrow$} & \textbf{R $\uparrow$} & \textbf{R(R) $\uparrow$} & & \\
\specialrule{1.2pt}{0pt}{0pt}
\textbf{DUSt3R} & 0.147 & 0.170 & 0.523 & 0.636 & 0.602 & 0.818 & 9.96 & 0.114 \\
\textbf{VGGT} & \textbf{\textcolor{blue}{0.579}} & \textbf{\textcolor{red}{0.647}} & 0.807 & \textbf{\textcolor{red}{0.954}} & 0.818 & \textbf{\textcolor{red}{0.965}} & \textbf{\textcolor{red}{3.31}} & 0.088 \\
\textbf{iFusion} & \textbf{\textcolor{red}{0.625}} & \textbf{\textcolor{blue}{0.625}} & \textbf{\textcolor{red}{0.875}} & \textbf{\textcolor{blue}{0.910}} & \textbf{\textcolor{red}{0.921}} & \textbf{\textcolor{blue}{0.954}} & \textbf{\textcolor{blue}{3.65}} & \textbf{\textcolor{red}{0.044}} \\
\textbf{Ours} & 0.489 & 0.500 & \textbf{\textcolor{blue}{0.818}} & 0.841 & \textbf{\textcolor{blue}{0.864}} & 0.898 & 5.07 & \textbf{\textcolor{blue}{0.072}} \\
\specialrule{1.5pt}{0pt}{0pt}
\end{tabular}
\end{adjustbox}
\label{tab:eval_results_unseen}
\vspace{-0.5em}
\end{table}

\begin{table}[t]
\centering
\caption{
\textbf{Multi-view joint reasoning.} We evaluate our framework for multi-view estimation, reporting recall at thresholds of $15^\circ$ and $30^\circ$. 
To analyze the contributions of each stage, we include ablation results: without Stage 1 (score-based initialization), without Stage 2 (Zero123 refinement), and using both Stage 1 and Stage 2.}
\vspace{-1.0em}
\renewcommand{\arraystretch}{1.2} 
    \begin{adjustbox}{width=\linewidth} 
    \begin{tabular}{l l c c c c c c c}
    \specialrule{1.5pt}{0pt}{0pt}
    & \textbf{\# of Images} & \textbf{2} & \textbf{3} & \textbf{4} & \textbf{5} & \textbf{6} & \textbf{7} & \textbf{8} \\
    \specialrule{1.2pt}{0pt}{0pt}
    \multirow{3}{*}{\textbf{R@15}}
    & \textbf{w/o Stage 1} & 0.200  & 0.103  &  0.065 &  0.075 &  0.074 & 0.071  & 0.025  \\
    & \textbf{w/o Stage 2} & 0.280 & 0.230 & 0.292 & 0.294 & 0.295 & 0.299 & 0.289 \\
    & \textbf{Stage 1 \& 2}  & \textbf{0.540} & \textbf{0.513} & \textbf{0.568} & \textbf{0.589} & \textbf{0.616} & \textbf{0.662} & \textbf{0.643} \\
    \hline
    \multirow{3}{*}{\textbf{R@30}} 
    & \textbf{w/o Stage 1} & 0.230 & 0.137 & 0.112 & 0.141 & 0.119 & 0.114 & 0.050 \\
    & \textbf{w/o Stage 2} & 0.450 & 0.443 & 0.507 & 0.528 & 0.544 & 0.550 & 0.521 \\
    & \textbf{Stage 1 \& 2} & \textbf{0.580} & \textbf{0.583} & \textbf{0.662} & \textbf{0.690} & \textbf{0.727} & \textbf{0.767} & \textbf{0.786}\\
    \specialrule{1.5pt}{0pt}{0pt}
    \end{tabular}
    \end{adjustbox}
\label{tab:multi_view}
\vspace{-0.5em}
\end{table}

\begin{table}[t]
\centering
\caption{\textbf{Ablation on different modeling approaches.} We compares score-based modeling and energy-based modeling on the GSO dataset with $10$ objects.}
\vspace{-1.0em}
\renewcommand{\arraystretch}{1.2}
\begin{adjustbox}{width=\linewidth}
\begin{tabular}{l c c c c c c c c}
\specialrule{1.5pt}{0pt}{0pt}
{\textbf{Method}} & {\textbf{R@5}} & {\textbf{R@15}} & {\textbf{R@30}} & {\textbf{SR@5}} & {\textbf{SR@15}} & {\textbf{SR@30}} & {\textbf{Rot.}} & {\textbf{Trans.}} \\
\specialrule{1.2pt}{0pt}{0pt}
\textbf{Score} & \textbf{0.700} & \textbf{0.963} & \textbf{0.963} & \textbf{0.631} & \textbf{0.900}  & \textbf{0.914} & \textbf{3.12} & 0.051 \\
\textbf{Energy} & 0.563 & 0.850 & 0.888 & 0.456 & 0.726 & 0.778 & 4.25 & \textbf{0.044} \\
\specialrule{1.5pt}{0pt}{0pt}
\end{tabular}
\end{adjustbox}
\label{tab:2methods_performance}
\vspace{-1.5em}
\end{table}



\vspace{-0.5em}
\subsection{Qualitative Results}
\vspace{-0.5em}

Fig.~\ref{fig:qualitative_result}~(a) presents qualitative comparisons on the synthetic GSO dataset. Our method consistently converges to correct poses, whereas iFusion often fails due to entrapment in local minima. Fig.~\ref{fig:qualitative_result}~(b) illustrates results on the real-world HOPEv2 dataset, where geometric symmetry introduces ambiguity. In these challenging cases, iFusion frequently converges to incorrect symmetric views that satisfy local optimality but fail to capture the true pose. In contrast, our method resolves these ambiguities, validating that our learned score function effectively guides the optimization away from suboptimal local minima and achieves robust convergence even with limited pose initialization.


\vspace{-0.7em}
\subsection{Ablation Study}
\label{sec:ablation_study}
\vspace{-0.5em}
\noindent\textbf{Evaluation under Various Numbers of Initializations.}\hspace{1em} We evaluate the robustness of our method under different numbers of initializations and compare it with iFusion, as shown in Fig.~\ref{fig:num_samples_recall}. Under a $30^\circ$ rotation threshold, our method achieves similar recall with only two initial poses, whereas iFusion requires eight. This empirical result clearly demonstrates the sample efficiency and robustness of our method, particularly in low-sample regimes.


\noindent\textbf{Ablation on Two Stage Design.}\hspace{1em}
To validate our two-stage framework, we evaluate the contribution of each stage on the multi-view pose estimation task. For each configuration, the number of input views varies from two to eight. As shown in Table~\ref{tab:multi_view}, removing Stage 1 leads to a significant performance degradation. This occurs since the solution space grows exponentially with the number of views, rendering optimization highly sensitive to initialization. This confirms that random sampling is ineffective and that the score model is essential for guiding optimization toward promising regions in the multi-view estimation task. Moreover, the incorporation of Stage 2 refinement consistently improves recall, substantiating the effectiveness of using the pretrained Zero123 as an energy-based refinement module.


\noindent\textbf{Comparison of Score-Based \& Energy-Based Methods.}\hspace{1em}
To assess the effectiveness of our score-based design, we compare it with the energy-based formulation introduced in Appendix~\ref{sec:derivation_details}. Both formulations aim to guide the pose update toward the correct pose. The results in Table~\ref{tab:2methods_performance} show that the score-based approach outperforms the energy-based approach. We attribute this to the fact that the score-based method predicts the score directly, whereas the energy-based model learns an energy function whose gradient is an indirect approximation of the score for pose updates.


\vspace{-1em}
\section{Conclusion}
\vspace{-0.8em}
\label{sec:conclusion_and_limitations}
In this work, we introduced a novel perspective on two-view camera pose estimation by analyzing it through the landscape perspective. We provided clear visualizations of the MSE landscape of the Zero123 model, and highlighted the local minima issues encountered by iFusion. Unlike previous methods that rely heavily on dense sampling, our approach leverages a learned score model to reshape the optimization dynamics, effectively guiding pose estimates toward the ground-truth and mitigating the impact of poor local minima. By incorporating this insight into a two-stage optimization scheme, our method achieves performance on par with with state-of-the-art methods while requiring fewer samples and significantly reducing inference time. These results demonstrate that a landscape-aware formulation not only enhances robustness to initialization but also paves the way for more sample-efficient and gradient-driven approaches to pose estimation.

\section{Acknowledgement}
The authors gratefully acknowledge the support from the National Science and Technology Council (NSTC) in Taiwan under grant numbers NSTC 114-2221-E-002-069-MY3, NSTC 113-2221-E-002-212-MY3, and NSTC 114-2218-E-A49-026, as well as the support from the Academia Sinica Scholar Award (ASSA) under grant number AS-ASSA-115-02, NTU Artificial Intelligence Center of Research Excellence, and Taiwan Centers of Excellence in Artificial Intelligence. This research was also supported by the NVIDIA Academic Grant Program. The authors would also like to express their appreciation for the donation of the GPUs from NVIDIA Corporation and NVIDIA AI Technology Center (NVAITC) used in this work. Furthermore, the authors extend their gratitude to the National Center for High-Performance Computing (NCHC) for providing computational and storage resources. The authors also thank the NVIDIA Taipei-1 supercomputer for providing essential computing resources.

\bibliographystyle{ieeenat_fullname}
\bibliography{main}

\clearpage
\appendix

\section*{Appendix}
This appendix provides supplementary materials to support the main manuscript. 
Section~\ref{sec:derivation_details} presents detailed derivations that support the theoretical foundations of our approach.  
Section~\ref{sec:implementation_details} outlines implementation specifics, including training settings, hyperparameters, dataset construction, and other relevant details.
Section~\ref{sec:additional_experimental_details} provides detailed experimental settings, extended results, and additional ablation studies.
Section~\ref{sec:visualization} presents qualitative results on camera pose estimation, as well as additional visualizations of optimization landscapes, trajectories, and score fields, offering insights into how these factors influence the pose estimation process.  
Section~\ref{sec:limitations} discusses limitations and potential directions for future work.

\vspace{-0.5em}
\section{Derivation Details}
\label{sec:derivation_details}
\vspace{-0.5em}
In this section, we provide detailed derivations and clarifications of the mathematical formulations presented in the manuscript. Section~\ref{sec:list_of_symbols} introduces the symbols and notations used throughout.
Section~\ref{sec:energy-based_modeling} introduces an energy-based formulation as an alternative to our score-based approach.
Section~\ref{sec:closed_form_expression} derives the closed-form training objectives corresponding to the learning approaches discussed in Section~\ref{sec:methodology}. Section~\ref{sec:proofs} presents the detailed proofs and theoretical analyses included in the manuscript.

\subsection{List of Symbols}
\label{sec:list_of_symbols}
This section summarized the symbols and notations used throughout the main manuscript and appendix. Table~\ref{tab:symbols} organizes them together with brief description.

\begin{table*}[ht]
\centering
\footnotesize
\caption{List of symbols and their corresponding descriptions}
\vspace{-1em}
\renewcommand{\arraystretch}{1.0}
\resizebox{1.0\textwidth}{!}{
\begin{tabular}{l| p{9cm}}
    \specialrule{1.2pt}{0pt}{0pt}
    \textbf{Symbol} & \textbf{Description} \\
    \specialrule{1.2pt}{0pt}{0pt}
    $\lVert u\rVert=\sqrt{u^T u}=\sqrt{\sum_i u_i^2}$ & Euclidean norm of vectors $u$. \\
    $\mathbf{I}_d$ & $d\times d$ identity matrix. \\
    $\mathbf{x}\in \mathbb{R}^d$ & Data sample with dimension $d$. \\
    $\tilde{\mathbf{x}}\in \mathbb{R}^d$ & Perturbed data sample. \\
    $\{\mathbf{x}^{(i)}\}_{i=1}^N$ & A dataset consisting of $N$ i.i.d. samples. \\
    $\{(\mathbf{x}^{(i)}, \mathbf{y}^{(i)})\}_{i=1}^N$ & The dataset consists of $N$ pairs, where each  pair includes a relative camera pose $\mathbf{x}^{(i)}$ condition on an image pair $\mathbf{y}^{(i)}$. \\
    $(\Theta, \Phi, \rho)$ & Spherical coordinates representing camera pose. \\
    \midrule
    $\nabla_\mathbf{x} f(\mathbf{x})$ & Gradient of function $f$ with respect to $\mathbf{x}$. \\
    $\mathcal{E}(\mathbf{x})$ & Energy function. \\
    $s (\mathbf{x})$ & Score function. \\
    $\delta(\mathbf{x})$ & Dirac-delta function. \\
    $p_{\text{data}}(\mathbf{x})$ & True underlying distribution of data samples $\mathbf{x}$. \\
    $p_\sigma(\tilde{\mathbf{x}} \mid \mathbf{x}) = \mathcal{N}(\tilde{\mathbf{x}} \mid \mathbf{x},\sigma^2\mathbf{I}_d)$ & Isotropic Gaussian kernel, where $\sigma^2$ denotes the variance.\\
    $p(\mathbf{x}\mid \mathbf{y})$ & Conditional distribution of the camera pose $\mathbf{x}$ given the image pair $\mathbf{y}$.\\
    \midrule
    $SE(3)$ & The Lie group representing 3D rigid body transformations. \\
    $\mathfrak{se}(3)$ & The Lie algebra associated with $SE(3)$. \\
    $T \in SE(3)$ & Transformation matrix between two coordinate systems. \\
    $\xi \in \mathfrak{se}(3)$ & Twist coordinates in the Lie algebra $\mathfrak{se}(3)$. \\
    $\text{Exp}(\cdot)$ & The exponential map from $\mathfrak{se}(3)$ to $SE(3)$.\\
    \specialrule{1.2pt}{0pt}{0pt}
\end{tabular}
}
\label{tab:symbols}
\vspace{-1.em}
\end{table*}

\subsection{Energy-based Modeling}
\label{sec:energy-based_modeling}
To further validate our design choice, we compare the score-based formulation with an energy-based alternative. Instead of explicitly learning the score function, we train an energy network $\mathcal{E}_\theta(\tilde{\mathbf{x}}, \mathbf{y})$ parameterized by $\theta$ to model the data distribution through the following objective:
\begin{equation}
\scalebox{0.96}{$\mathcal{L}(\theta) = \frac{1}{2} \mathbb{E}_{\mathbf{x}, \mathbf{y}} \mathbb{E}_{\tilde{\mathbf{x}}\sim \mathrm{U}}  \left\| \mathcal{E}_\theta(\tilde{\mathbf{x}}, \mathbf{y}) + \log p_\sigma(\tilde{\mathbf{x}} \mid \mathbf{x}, \mathbf{y}) \right\|_2^2.$}
\label{eq:train_obj_energy}
\end{equation}
After training, the gradient of the learned energy function with respect to $\tilde{\mathbf{x}}$ yields the corresponding score, which can be integrated into our two-stage optimization framework. A comparison of this energy-based approach with our proposed score-based design highlights the efficiency and stability advantages of the latter, as it directly learns the score field without requiring differentiation of the energy function. The quantitative results comparing these two modeling approaches are shown in Table~\ref{tab:2methods_performance}.

\subsection{Closed-Form Expression}
\label{sec:closed_form_expression}
Closed-form expressions for the score and energy targets in Eq.~(\ref{eq:train_obj_score}) and Eq.~(\ref{eq:train_obj_energy}) are derived using a Gaussian perturbation kernel:
$p_\sigma(\tilde{\mathbf{x}} \mid \mathbf{x}) = \frac{1}{(2\pi \sigma^2)^{d/2}} \exp\left(-\frac{\lVert \tilde{\mathbf{x}} - \mathbf{x} \rVert^2}{2\sigma^2}\right)$, where $d$ denotes the dimensionality of the data space and $\sigma$ denotes the standard deviation. The corresponding log-density is given by
\begin{equation}
    \scalebox{1.0}{$\log p_\sigma(\tilde{\mathbf{x}} \mid \mathbf{x}) = C - \frac{\lVert \tilde{\mathbf{x}} - \mathbf{x} \rVert^2}{2\sigma^2},$}
    \label{eq:logp_expression}
\end{equation}
\noindent where $C = -\frac{d}{2} \log(2\pi \sigma^2)$ is a contant independent of $\tilde{\mathbf{x}}$.

\noindent\textbf{Score-based objective.}\hspace{1em}
Differentiation of Eq.~(\ref{eq:logp_expression}) with respect to $\tilde{\mathbf{x}}$ yields the score function
\begin{equation}
    \nabla_{\tilde{\mathbf{x}}} \log p_\sigma(\tilde{\mathbf{x}} \mid \mathbf{x}) = -\frac{\tilde{\mathbf{x}} - \mathbf{x}}{\sigma^2}.
    \label{eq:score_expression}
\end{equation}
Substituting Eq.~(\ref{eq:score_expression}) into the training objective in Eq.~(\ref{eq:train_obj_score}) yields the explicit expression for the training loss:
\begin{equation}
    \scalebox{1.0}{$\mathcal{L}(\theta) = \frac{1}{2} \mathbb{E}_{\mathbf{x}, \mathbf{y}} \mathbb{E}_{\tilde{\mathbf{x}}\sim \mathrm{U}}  \left\| s_\theta(\tilde{\mathbf{x}}, \mathbf{y}) + \frac{\tilde{\mathbf{x}} - \mathbf{x}}{\sigma^2} \right\|^2.$}
\end{equation}
\noindent where $s_\theta(\tilde{\mathbf{x}}, \mathbf{y})$ denotes the score network.

\noindent\textbf{Energy-based objective.}\hspace{1em}
To construct the energy target used in energy-based modeling, we first observe that the constant term $C$ in Eq.~(\ref{eq:logp_expression}) is independent of the noised input $\tilde{\mathbf{x}}$ and can therefore be omitted during training. Consequently, the objective depends solely on the quadratic term. Substituting this expression into Eq.~(\ref{eq:train_obj_energy}) yields the energy-based training objective
\begin{equation}
    \scalebox{1.0}{$\mathcal{L}(\theta) = \frac{1}{2} \mathbb{E}_{\mathbf{x}, \mathbf{y}} \mathbb{E}_{\tilde{\mathbf{x}}\sim \mathrm{U}}  \left\| \mathcal{E}_\theta(\tilde{\mathbf{x}}, \mathbf{y}) - \frac{\lVert \tilde{\mathbf{x}} - \mathbf{x} \rVert^2}{2\sigma^2}\right\|^2,$}
\end{equation}
\noindent where $\mathcal{E}_\theta(\tilde{\mathbf{x}}, \mathbf{y})$ denotes the energy network.

\subsection{Proofs}
\label{sec:proofs}
\begin{proof}[Proof of Proposition~1.]
To derive the optimal score function $s_\theta (\tilde{\mathbf{x}}, \mathbf{y})$, we reformulate the objective in Eq.~(\ref{eq:DSM_obj}) by rearranging the expectations over $\mathbf{y}$, $\tilde{\mathbf{x}}$, and $\mathbf{x}$, yielding
\begin{equation}
    \scalebox{1.0}{$\mathcal{L}(\theta) = \frac{1}{2} \mathbb{E}_{\mathbf{y}} \mathbb{E}_{\tilde{\mathbf{x}}} \mathbb{E}_{\mathbf{x}}  \left\| s_\theta(\tilde{\mathbf{x}}, \mathbf{y}) - \nabla_{\tilde{\mathbf{x}}} \log p_\sigma(\tilde{\mathbf{x}} \mid \mathbf{x}) \right\|^2,$}
\end{equation}
\noindent where the expectations are taken over $\mathbf{y} \sim p(\mathbf{y}), \tilde{\mathbf{x}} \sim p(\tilde{\mathbf{x}} \mid \mathbf{y})$, and $\mathbf{x} \sim p(\mathbf{x} \mid \mathbf{y}, \tilde{\mathbf{x}})$. 
For fixed $(\tilde{\mathbf{x}}, \mathbf{y})$, the inner expectation defines a quadratic loss:
\begin{equation}
    \mathcal{L}(s)=\mathbb{E}_{\mathbf{x}\sim p(\mathbf{x}\mid\mathbf{y}, \tilde{\mathbf{x}})}  \left\| s - \nabla_{\tilde{\mathbf{x}}} \log p_\sigma(\tilde{\mathbf{x}} \mid \mathbf{x}) \right\|^2,
    \label{eq:quadratic_loss}
\end{equation}
where $s=s_\theta(\tilde{\mathbf{x}},\mathbf{y})$ is treated as the optimization variable. By minimizing the quadratic loss in Eq.~(\ref{eq:quadratic_loss}) and setting $\nabla_s\mathcal{L}(s)=0$, we obtain the optimal solution:
\begin{equation}
    s^\star(\tilde{\mathbf{x}}, \mathbf{y})=\mathbb{E}_{\mathbf{x}\sim p(\mathbf{x}\mid\mathbf{y}, \tilde{\mathbf{x}})}[\nabla_{\tilde{\mathbf{x}}} \log p_\sigma(\tilde{\mathbf{x}} \mid \mathbf{x})].
    \label{eq:opt_s}
\end{equation}
Substituting the closed-form expression of $p_\sigma(\tilde{\mathbf{x}}\mid\mathbf{x})$ from Eq.~(\ref{eq:score_expression}) into Eq.~(\ref{eq:opt_s}) yields
\begin{align}
    s^\star(\tilde{\mathbf{x}},\mathbf{y})
    &=\mathbb{E}_{\mathbf{x}\sim p(\mathbf{x}\mid\mathbf{y},\tilde{\mathbf{x}})}[\nabla_{\tilde{\mathbf{x}}}\log p_\sigma(\tilde{\mathbf{x}}\mid\mathbf{x})] \notag \\
    &=\int \nabla_{\tilde{\mathbf{x}}}\log    p_\sigma(\tilde{\mathbf{x}}\mid\mathbf{x})\,p(\mathbf{x}\mid\mathbf{y},\tilde{\mathbf{x}})\,d\mathbf{x} \notag \\
    &=\frac{\int(\mathbf{x}-\tilde{\mathbf{x}})\,p(\tilde {\mathbf{x}}\mid\mathbf{x})\,p(\mathbf{x}\mid\mathbf{y})\,d\mathbf{x}}{\sigma^2\int p(\tilde{\mathbf{x}}\mid\mathbf{x})\,p(\mathbf{x}\mid\mathbf{y})\,d\mathbf{x}} \,.
    \label{eq:optimal_score_expression}
\end{align}
In the last step in Eq.~(\ref{eq:optimal_score_expression}), we used the fact that the perturbation $\tilde{\mathbf{x}}$ depends only on the original variable $\mathbf{x}$ and is independent of $\mathbf{y}$. Formally, this means
$
p(\tilde{\mathbf{x}} \mid \mathbf{y}, \mathbf{x}) = p(\tilde{\mathbf{x}} \mid \mathbf{x}),
$
which allows us to factor $p(\mathbf{x} \mid \mathbf{y}, \tilde{\mathbf{x}})$ as $p(\tilde{\mathbf{x}} \mid \mathbf{x}) \, p(\mathbf{x} \mid \mathbf{y})$ in the integral.
Hence, the optimal score function of Eq.~(\ref{eq:DSM_obj}) takes the form above, as stated in Proposition~1. \qedhere
\end{proof}

\begin{proof}[Proof of Lemma~1.]
Since the perturbation variable $\tilde{\mathbf{x}}$ is drawn from a uniform distribution $\mathrm{U}$ that is independent of $\mathbf{x}$, we have $p(\mathbf{x}\mid\mathbf{y}, \tilde{\mathbf{x}})=p(\mathbf{x}\mid\mathbf{y})$. Substituting this into Eq.~(\ref{eq:opt_s}) from Proposition~1, we obtain
\begin{equation}
    s_\mathrm{U}^\star(\tilde{\mathbf{x}}, \mathbf{y})=\mathbb{E}_{\mathbf{x}\sim p(\mathbf{x}\mid\mathbf{y})}\left[\nabla_{\tilde{\mathbf{x}}} \log p_\sigma(\tilde{\mathbf{x}} \mid \mathbf{x})\right].
    \label{eq:opt_s_uniform}
\end{equation}
Using the analytic form of $p_\sigma(\tilde{\mathbf{x}} \mid \mathbf{x})$ from Eq.~(\ref{eq:score_expression}), Eq.~(\ref{eq:opt_s_uniform}) can be expressed in the following explicit integral form:
\begin{equation}
    s_\mathrm{U}^\star(\tilde{\mathbf{x}},\mathbf{y})=\frac{1}{\sigma^2}\int (\mathbf{x}-\tilde {\mathbf{x}})\,p(\mathbf{x}\mid\mathbf{y})\,d\mathbf{x}.
\end{equation}
This concludes the derivation and verifies Lemma 1. \qedhere
\end{proof}

\begin{proof}[Proof of Lemma~2.]
By Bayes' rule and the conditional independence of $\tilde{\mathbf{x}}$ from $\mathbf{y}$ given $\mathbf{x}$, the posterior distribution in Eq.~(\ref{eq:opt_s}) can be written in the following form
\begin{equation}
    p(\mathbf{x}\mid\mathbf{y},\tilde{\mathbf{x}})=\frac{p(\tilde {\mathbf{x}}\mid\mathbf{x})\,p(\mathbf{x}\mid\mathbf{y})}{\int p(\tilde {\mathbf{x}}\mid\mathbf{x})\,p(\mathbf{x}\mid\mathbf{y})\,d\mathbf{x}}.
    \label{eq:bayes_posterior}
\end{equation}
The posterior depends on both the conditional prior $p(\mathbf{x}\mid\mathbf{y})$ and the likelihood $p(\tilde {\mathbf{x}}\mid\mathbf{x})$. 
In general, if the conditional prior $p(\mathbf{x}\mid \mathbf{y})$ has support on multiple values of $\mathbf{x}$, the posterior distribution in Eq.~(\ref{eq:bayes_posterior}) differs from the prior.
Consequently, the optimal score functions in Eqs.~(\ref{eq:opt_s}) and (\ref{eq:opt_s_uniform}) are generally different: $s^\star(\tilde{\mathbf{x}},\mathbf{y}) \neq s_\mathrm{U}^\star(\tilde{\mathbf{x}},\mathbf{y})$.
However, in the special case where each conditional prior collapses to a Dirac delta, $p(\mathbf{x}\mid\mathbf{y}^{(i)})=\delta(\mathbf{x}-\mathbf{x}^{(i)})$, each conditional image pair $\mathbf{y}^{(i)}$ corresponds to a single ground-truth pose $\mathbf{x}^{(i)}$. Substituting the Dirac delta function into Eq.~(\ref{eq:bayes_posterior}) yields
\begin{align}
p(\mathbf{x}\mid\mathbf{y}^{(i)},\tilde{\mathbf{x}}) &= \frac{p(\tilde {\mathbf{x}}\mid\mathbf{x})\,\delta(\mathbf{x}-\mathbf{x}^{(i)})}
{\int p(\tilde{\mathbf{x}}\mid\mathbf{x})\,\delta(\mathbf{x}-\mathbf{x}^{(i)})\,d\mathbf{x}}\notag\\
&= \frac{p(\tilde {\mathbf{x}}\mid\mathbf{x})\,\delta(\mathbf{x}-\mathbf{x}^{(i)})}
{p(\tilde{\mathbf{x}}\mid\mathbf{x}^{(i)})} \notag \\
&= \delta(\mathbf{x}-\mathbf{x}^{(i)}).
\end{align}
In this case, the posterior coincides with the prior, i.e., $p(\mathbf{x} \mid \mathbf{y}^{(i)})=p(\mathbf{x} \mid \mathbf{y}^{(i)}, \tilde{\mathbf{x}})$. As a result, the optimal score functions are identical: $s^\star(\tilde{\mathbf{x}}, \mathbf{y}) = s_\mathrm{U}^\star(\tilde{\mathbf{x}}, \mathbf{y})$. \qedhere 
\end{proof}

\begin{proof}[Proof of Eq.~(\ref{eq:decay_converge}).]
We assume the learned score function approximates the true score $s_\theta(\tilde{\mathbf{x}},\mathbf{y}) = (\mathbf{x}-\tilde{\mathbf{x}})/\sigma^2$ for each $(\mathbf{x}, \mathbf{y})$ in the dataset. For clarity in the following analysis, we denote the ground-truth pose associated with the conditional image pair $\mathbf{y}$ as $\mathbf{x}_{\text{gt}}$. In our experiments, we set $\sigma=1$. Substituting these into Eq.~(\ref{eq:update_function}) yields the update equation:
\begin{equation}
    \tilde{\mathbf{x}}_t = \tilde{\mathbf{x}}_{t-1} + \alpha (\mathbf{x}_{\mathrm{gt}}-\tilde{\mathbf{x}}_{t-1}) + G\mathbf{z}_t,
    \quad \mathbf{z}_t \sim \mathcal{N}(0, \mathbf{I}_3).
\label{eq:linear_update}
\end{equation}
Starting from the initial point $\tilde{\mathbf{x}}_0$, the iterative formula in Eq.~(\ref{eq:linear_update}) updates the noisy prediction $\tilde{\mathbf{x}}_{t-1}$ to $\tilde{\mathbf{x}}_{t}$.
Taking the expectation on both sides of Eq.~(\ref{eq:linear_update}) gives
\begin{equation}
    \mathbb{E}[\tilde{\mathbf{x}}_{t}]=(1 - \alpha)\, \mathbb{E}[\tilde{\mathbf{x}}_{t-1}] + \alpha\, \mathbf{x}_{\mathrm{gt}},
\label{eq:linear_update_expec}
\end{equation}
where the noise term vanishes since $\mathbb{E}[\mathbf{z}_t] = \mathbf{0}$.
Eq.~(\ref{eq:linear_update_expec}) can be rewritten as the following recurrence:
\begin{equation}
    \delta_t = (1 - \alpha)\, \delta_{t-1},
    \label{eq:recurrence}
\end{equation}
where $\delta_t = \mathbb{E}[\tilde{\mathbf{x}}_{t} - \mathbf{x}_{\mathrm{gt}}]$. Since both $\tilde{\mathbf{x}}_0$ and $\mathbf{x}_{\mathrm{gt}}$ are fixed, the expectation is redundant at $t=0$, giving $\delta_0 = \tilde{\mathbf{x}}_0 - \mathbf{x}_{\mathrm{gt}}$.

\noindent Solving Eq.~(\ref{eq:recurrence}) gives $\delta_t = (1 - \alpha)^t \delta_0$. Substituting the definition of $\delta_t$ and taking norms on both sides then yields
\begin{equation}
    \lVert\mathbb{E}[\tilde{\mathbf{x}}_t - \mathbf{x}_{\mathrm{gt}}]\rVert = M\,(1 - \alpha)^t,
    \label{eq:converge_expect}
\end{equation}
where $M = \lVert \tilde{\mathbf{x}}_0 - \mathbf{x}_{\mathrm{gt}} \rVert$ denotes the initial prediction error and provides an upper bound on the expected distance from the ground-truth pose.
Next, we compute the variance of the noised prediction $\tilde{\mathbf{x}}_t$. By taking the variance on both sides of the iterative update in Eq.~(\ref{eq:linear_update}), we obtain
\begin{align}
    \mathrm{Var}[\tilde{\mathbf{x}}_{t}]
    &= \mathrm{Var}[(1-\alpha)\,\tilde{\mathbf{x}}_{t-1} + G\, \mathbf{z}_t] \notag\\
    &= (1-\alpha)^2 \, \mathrm{Var}[\tilde{\mathbf{x}}_{t-1}] + G \, \mathrm{Var}[\mathbf{z}_t] \, G^\top.
\end{align}
Since $\mathbf{z}_t \sim \mathcal{N}(0, \mathbf{I}_3)$, we have $\mathrm{Var}[\mathbf{z}_t] = \mathbf{I}_3$. As the multiplicative factor $(1-\alpha)^2<1$, the variance recursion is guaranteed to converge. By setting $\mathrm{Var}[\tilde{\mathbf{x}}_{t}] = \mathrm{Var}[\tilde{\mathbf{x}}_{t-1}] = \Sigma$ and using the fact that $G$ is a diagonal matrix so that $G\,G^\top = G^2$, we obtain the equation $\Sigma = (1-\alpha)^2 \Sigma + G^2$. By solving the equation for $\Sigma$, the solution is
\begin{equation}
    \Sigma = \frac{G^2}{1 - (1-\alpha)^2} = \frac{G^2}{2\alpha - \alpha^2} \approx \frac{G^2}{2\alpha}.
    \label{eq:converge_variance}
\end{equation}
With the expectation result in Eq.~(\ref{eq:converge_expect}) and the variance result in Eq.~(\ref{eq:converge_variance}), the dynamics of the iterative update are fully characterized. This completes the derivation and thus concludes the proof of Eq.~(\ref{eq:decay_converge}). \qedhere
\end{proof}

\noindent Eq.~(\ref{eq:decay_converge}) indicates that, under an accurate score approximation, the distance between the predicted pose and the ground-truth pose decays exponentially, while the variance is governed by the coordinate-wise noise scales $\gamma_i$. These results provide a theoretical justification for the design of our score-based first-stage optimization.

\section{Implementation Details}
\label{sec:implementation_details}
In this section, we provide implementation details to support reproducibility.
Section~\ref{sec:coordinate_system} introduces the spherical coordinate system used to represent camera poses in our work.
Section~\ref{sec:dataset_construction} describes the dataset construction and rendering procedures.
Section~\ref{sec:training_and_evaluation} details the training setup, model architecture, and our proposed two-stage optimization strategy.
Section~\ref{sec:computation_resources} summarizes the computation resources employed in our experiments..  

\begin{figure}[t]
  \centering
  \includegraphics[width=0.9\linewidth]{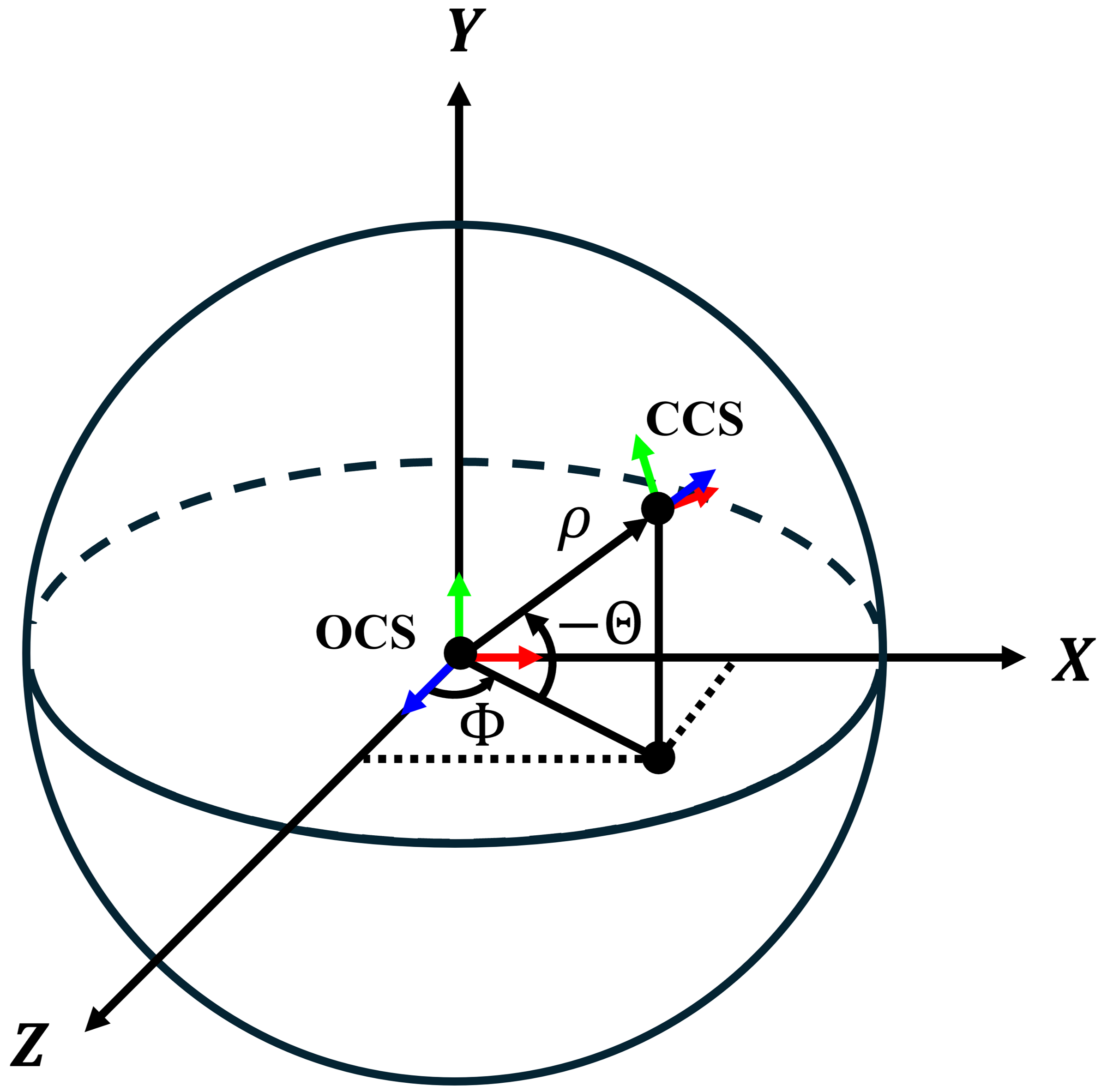}
  \caption{\textbf{Spherical Coordinate System.} \textbf{CCS} denotes camera coordinate system and \textbf{OCS} denotes object coordinate system.
  }
  \label{fig:spherical_coordinate}
\end{figure}

\subsection{Coordinate System}
\label{sec:coordinate_system}
Following iFusion~\citep{wu2023ifusion}, we represent camera positions and their relative transformations using a spherical coordinate system centered at the object’s origin, as illustrated in Fig.~\ref{fig:spherical_coordinate}. The coordinates $\Theta$, $\Phi$, and $\rho$ denote the vertical angle from the equatorial plane, the horizontal angle within that plane, and the radial distance from the origin, respectively. Since we assume an object-centric camera pose, the extrinsic matrix between the camera coordinate system (CCS) and the object coordinate system (OCS) is uniquely determined by the camera's location in this spherical coordinate system. During training, given two images captured from different viewpoints with camera positions $(\Theta_1, \Phi_1, \rho_1)$ and $(\Theta_2, \Phi_2, \rho_2)$, we define their relative transformation as the difference vector $(\Theta_2 - \Theta_1, \Phi_2 - \Phi_1, \rho_2 - \rho_1)$. For clarity and ease of interpretation, we refer to $\Theta$ as latitude and $\Phi$ as longitude throughout this paper to more intuitively represent the camera pose.

\subsection{Dataset Construction}
\label{sec:dataset_construction}
\textbf{GSO and OO3D.}\hspace{1em} We construct our training and testing datasets using 3D models from the GSO and OO3D datasets, following the data preparation pipeline of iFusion. Both datasets use the same 70 objects as in the iFusion dataset. To assess the generalization capability of our framework, we also selected 10 additional objects from the GSO dataset that were not used during training. Each object was normalized to fit within a unit cube by scaling its maximum side length to one. Rendering was performed using Pyrender~\citep{pyrender} with a perspective camera whose field of view (FoV) was set to $49.1^\circ$. All images were rendered at a resolution of $512 \times 512$ pixels with transparent backgrounds. Camera positions were defined using the spherical coordinate system introduced in Section~\ref{sec:coordinate_system}.

The training data was generated by uniformly sampling camera viewpoints on the unit sphere. The latitude $\Theta$ were sampled from the range $[-30^\circ, 30^\circ]$ in $7.5^\circ$ increments, and the longitude $\Phi$ from $[0^\circ, 360^\circ]$ in $15^\circ$ increments. The camera distance $\rho$ was uniformly sampled from the interval $[1.2, 2.0]$. This resulted in a total of $15120$ training images for each dataset. 
The testing viewpoints were randomly sampled from the same parameter ranges, with latitude $\Theta \in [-30^\circ, 30^\circ]$, longitude $\Phi \in [0^\circ, 360^\circ]$, and camera distance $\rho \in [1.2, 2.0]$. For each object, $8$ viewpoints were randomly selected, resulting in a total of $560$ testing images under this configuration.

\noindent \textbf{HOPEv2.}\hspace{1em} For the HOPEv2 dataset, which contains $28$ objects, we uniformly sampled 100 images per object from the original dataset to cover diverse viewpoints, resulting in a total of $2800$ training images. For testing, $4$ images per object were selected, yielding a total of $112$ testing images. To satisfy the object-centric assumption, each image was cropped around the object, and the corresponding ground-truth annotations were adjusted accordingly. The images were preprocessed using the object masks to isolate the objects from the background.

\subsection{Training and Evaluation}
\label{sec:training_and_evaluation}
\vspace{-0.5em}
\textbf{Model Architecture and Hyperparameters.}\hspace{1em} The score network is conditioned on an image pair and a relative noised pose. The reference and query images are first passed through a ResNet-50~\citep{DBLP:conf/cvpr/HeZRS16} backbone to extract 2048-dimensional feature vectors for each image. The conditioned pose is embedded using a sinusoidal positional encoding, and these features are concatenated to form the model input.
The concatenated input is then passed through a lightweight MLP with residual connections~\citep{DBLP:conf/cvpr/HeZRS16}, ReLU activations, and layer normalization~\citep{DBLP:journals/corr/BaKH16}. The number of hidden channels is set to $256$. The network outputs a three-dimensional vector corresponding to the conditional score of the data distribution. Through empirical tuning, we found that this architecture is sufficient to accurately learn the score function and effectively guide the perturbed pose toward the ground-truth pose.

Our implementation is based on PyTorch~\citep{DBLP:conf/nips/PaszkeGMLBCKLGA19}. For all experiments, we used the Adam~\citep{DBLP:journals/corr/KingmaB14} optimizer with $\beta_1=0.9$, $\beta_2=0.999$, a learning rate of $0.0001$, a batch size of $256$, and $60$ epochs. For each object, we consider all possible pairs of image. For each image pair, the noised pose is randomly sample from the uniform distribution $U=[-\frac{\pi}{3}, \frac{\pi}{3}]\times [0, 2\pi]\times [1.2, 2.0]\subset \mathbb{R}^3$. In the second stage of our two-stage optimization pipeline, the pose-conditioned diffusion model uses Zero123-XL as its 3D geometric prior. Comparison with other alternatives are provided in Section~\ref{sec:more_ablation_studies}.


\noindent \textbf{Two Stage Optimization.}\hspace{1em} After training, the learned score function can be used to guide arbitrary
initial poses toward higher-density regions of the pose distribution. The first-stage optimization is performed by iteratively applying the update rule defined in Eq.~(\ref{eq:update_function}). In practice, we set the hyperparameters to $\alpha = 0.1, \gamma_1=\gamma_3=0, \gamma_2=0.3$, and run the optimization for $50$ iterations.

\noindent In the second stage, following iFusion, the prediction is refined using the pretrained Zero123 model with an MSE loss. This involves computing the gradient with respect to the pose condition (i.e., the noised pose estimate),  followed by applying a gradient-based iterative solver for optimization. To ensure that the estimated pose remains on the $\mathrm{SE}(3)$ manifold during this process, we parameterize it as $T_{r \rightarrow q} = \text{Exp}(\xi)$, where $\xi \in \mathbb{R}^6$ denotes the twist vector in the Lie algebra~\citep{DBLP:journals/corr/abs-1812-01537} $\mathfrak{se}(3)$ associated with the Lie group $\mathrm{SE}(3)$. The exponential map $\text{Exp}(\cdot)$ converts the twist into a corresponding rigid body transformation in $SE(3)$.
The loss function is defined as:
\begin{equation}
    \scalebox{0.82}{$\hat \xi=\text{argmin}_{\xi\in \mathfrak{se}(3)}\mathcal{L}(I_q, (I_r, \text{Exp}(\xi)))+\mathcal{L}(I_r, (I_q, \text{Exp}(-\xi)))$,}
    \label{eq:iFusion_loss}
\end{equation}
where \scalebox{0.88}{$\mathcal{L}(I_q, (I_r, \text{Exp}(\xi))) = \mathbb{E}_{\mathbf{z}, \epsilon, t} \left[ \left\| \epsilon - \epsilon_\theta(\mathbf{z}_t, t, (I_r, \text{Exp}(\xi))) \right\|_2^2 \right]$}, $\epsilon$ is drawn from an isotropic Gaussian distribution, $t\in [0, 1, \ldots, 99]$, $\mathbf{z}$ denoting the latent representation of the query image $I_q$. 
The Adam optimizer is applied with a learning rate of $0.1$ for $50$ iterations. A \texttt{ReduceLROnPlateau} scheduler with a decay factor of $0.7$ and a patience of $10$ adaptively reduces the learning rate when the optimization plateaus. Each iteration corresponds to a diffusion timestep, obtained by linearly mapping the iteration index to the predefined diffusion time range. 


\subsection{Computation Resources}
\label{sec:computation_resources}
The training of our model was performed using a single NVIDIA RTX 6000 Ada GPUs, with 48\,GB of memory, together with an AMD EPYC 7313 16-core CPU. Training the score model on the 70-object GSO dataset (Table~\ref{tab:eval_results_myGSO_myOO3D}) takes approximately $1.5$ days.

\section{Additional Experimental Results}
\label{sec:additional_experimental_details}
In this section, we provide further experimental information to supplement the results reported in the main manuscript.
Section~\ref{sec:details_of_baseline_settings} describes the processing used for comparison with the baseline methods. Section~\ref{sec:analysis_of_inference_time} presents a comparison of inference time between iFusion and our method, both of which are gradient-based solver. Section~\ref{sec:more_ablation_studies} presents further ablation studies of our proposed approach.



\subsection{Details of Baseline Settings}
\label{sec:details_of_baseline_settings}
Our proposed method is compared against three baseline approaches: iFusion~\citep{wu2023ifusion}, DUSt3R~\citep{wang2024dust3r} and VGGT~\citep{DBLP:conf/cvpr/WangCKV0N25}. DUSt3R and VGGT estimates relative camera transformations from monocular image pairs, and the absolute scale of the translation cannot be determined without additional depth information. In our work, we follow the settings of Zero123 and iFusion, and learn a normalized representation of the object to predict a normalized camera scale, which is then refines in a second stage using Zero123. To ensure a fair comparison, we determine the optimal scale for these baseline methods. This ensures consistency with the assumed camera pose representation in our framework.

\begin{table*}[t]
\centering
\caption{\textbf{Evaluation results on the CO3Dv2 dataset.} For large-scale, scene-level evaluation, we compare the Zero123 gradient-based method on 29 sampled CO3Dv2 scenes. \textbf{\textcolor{red}{Red}} marks the best result, and \textbf{\textcolor{blue}{blue}} the second-best.}
\vspace{-1em}
\renewcommand{\arraystretch}{1.2} 
    \begin{adjustbox}{width=1.0\textwidth} 
    \begin{tabular}{l l c c c c c c c c c c c c c c}
    \specialrule{1.5pt}{0pt}{0pt}
    \multirow{2}{*}{\textbf{Dataset}} & \multirow{2}{*}{\textbf{Method}} & \multicolumn{2}{c}{\textbf{@5}} & \multicolumn{2}{c}{\textbf{@15}} & \multicolumn{2}{c}{\textbf{@30}} & \multicolumn{2}{c}{\textbf{@5}} & \multicolumn{2}{c}{\textbf{@15}} & \multicolumn{2}{c}{\textbf{@30}} & \multirow{2}{*}{\textbf{Rot. $\downarrow$}} & \multirow{2}{*}{\textbf{Trans. $\downarrow$}} \\
    \cmidrule(lr){3-4} \cmidrule(lr){5-6} \cmidrule(lr){7-8} \cmidrule(lr){9-10} \cmidrule(lr){11-12} \cmidrule(lr){13-14}
    & & \textbf{R $\uparrow$} & \textbf{R(R) $\uparrow$} & \textbf{R $\uparrow$} & \textbf{R(R) $\uparrow$} & \textbf{R $\uparrow$} & \textbf{R(R) $\uparrow$} & \textbf{SR $\uparrow$} & \textbf{SR(R) $\uparrow$} & \textbf{SR $\uparrow$} & \textbf{SR(R) $\uparrow$} & \textbf{SR $\uparrow$} & \textbf{SR(R) $\uparrow$} & & \\

    \specialrule{1.2pt}{0pt}{0pt}
    \multirow{3}{*}{\textbf{CO3Dv2}} & \textbf{ID-Pose}~\citep{cheng2023id} & 0.022 & 0.035 & 0.134 & 0.297 & 0.181 & 0.491 & 0.004 & 0.006 & 0.030 & 0.070 & 0.050 & 0.140 & 30.95 & 0.267 \\
    & \textbf{iFusion}~\citep{wu2023ifusion} & \textbf{\textcolor{red}{0.103}} & \textbf{\textcolor{red}{0.116}} & \textbf{\textcolor{blue}{0.392}} & \textbf{\textcolor{red}{0.522}} & \textbf{\textcolor{blue}{0.509}} & \textbf{\textcolor{blue}{0.711}} & \textbf{\textcolor{blue}{0.041}}  & \textbf{\textcolor{blue}{0.052}} & \textbf{\textcolor{blue}{0.176}} & \textbf{\textcolor{blue}{0.233}} & \textbf{\textcolor{blue}{0.237}} & \textbf{\textcolor{blue}{0.334}} & \textbf{\textcolor{red}{13.98}} & \textbf{\textcolor{blue}{0.160}} \\
    & \textbf{Ours} & \textbf{\textcolor{blue}{0.069}} & \textbf{\textcolor{blue}{0.078}} & \textbf{\textcolor{red}{0.401}} & \textbf{\textcolor{blue}{0.517}} & \textbf{\textcolor{red}{0.599}} & \textbf{\textcolor{red}{0.884}} &  \textbf{\textcolor{red}{0.054}}& \textbf{\textcolor{red}{0.071}} & \textbf{\textcolor{red}{0.344}} & \textbf{\textcolor{red}{0.469}} & \textbf{\textcolor{red}{0.567}} & \textbf{\textcolor{red}{0.838}} & \textbf{\textcolor{blue}{14.59}} & \textbf{\textcolor{red}{0.150}} \\
    \specialrule{1.5pt}{0pt}{0pt}
    \end{tabular}
    \end{adjustbox}
\vspace{-1.3em}
\label{tab:eval_results_CO3Dv2}
\end{table*}

\subsection{Analysis of Inference Time}
\label{sec:analysis_of_inference_time}
\begin{table}[t]
    \caption{\textbf{Comparison of inference time and performance.} We report recall (\textbf{R@30}) and inference time for different numbers of initial poses for each method. Our method achieves comparable performance while requiring substantially less inference time.}
    
    \label{tab:inference_time}
    \centering
    \resizebox{\columnwidth}{!}{%
    \begin{tabular}{c | c c c c c c c c}
        \toprule
        \textbf{Method} & \multicolumn{4}{c}{\textbf{iFusion}} & \multicolumn{4}{c}{\textbf{Ours}} \\
        \cmidrule(lr){1-5} \cmidrule(lr){6-9}
        \textbf{\# of Initial Poses} & \textbf{2} & \textbf{4} & \textbf{6} & \textbf{8} & \textbf{2} & \textbf{4} & \textbf{6} & \textbf{8} \\
        \midrule
        \textbf{R@30} & 0.661 & 0.817 & 0.893 & 0.907 & 0.901 & 0.913 & 0.943 & 0.938 \\
        \textbf{Time (s)} & 23.30 & 45.89 & 69.72 & 91.92 & 12.86 & 25.61 & 36.97 & 51.29 \\
        \bottomrule
    \end{tabular}
    }
\end{table}
Table~\ref{tab:inference_time} compares the inference time of iFusion and our proposed method across different numbers of initial poses. For comparable recall values, our method requires significantly less computation time due to the sample efficiency of our two-stage optimization. For instance, to achieve a recall near $0.90$, our approach takes only $12.86$ seconds with $2$ initial poses, whereas iFusion requires $91.92$ seconds with $8$ initial poses to achieve similar performance. Even for the same number of initial poses, our approach is faster because of its lightweight score model, compared to the heavy Zero123 diffusion UNet. These result demonstrates that our method is more robust to variations in the initial poses and achieves comparable performance with substantially fewer samples, resulting in a substantial reduction in runtime.

\subsection{Additional Quantitative Results}
Table~\ref{tab:eval_results_CO3Dv2} provides further comparisons with ID-Pose and iFusion on the large-scale CO3Dv2 3D category dataset. Training and testing data are sampled from the CO3Dv2 single-sequence subset, comprising 29 scenes, with viewpoints uniformly sampled for each scene. Our framework continues to produce strong results across all metrics, notably achieving the highest success rate among all methods.

\subsection{More Ablation Studies}
\label{sec:more_ablation_studies}
\begin{table}[t]
\centering
\caption{\textbf{Ablation on the noise scale $\gamma$.} The noise scale $\gamma$ controls the magnitude of the noise added during each iteration, which in turn affects the variance of the first-stage predictions. Here, we ablate the noise scale for the longitude dimension. Higher noise scales lead to higher recall, whereas lower noise scales yield higher success rates. \textbf{Bold} indicates the best result for each metric.}
\vspace{-0.5em}
\renewcommand{\arraystretch}{1.2}
\begin{adjustbox}{width=\linewidth}
\begin{tabular}{c l c c c c c c c c}
\specialrule{1.5pt}{0pt}{0pt}
& {} & {\textbf{R@5}} & {\textbf{R@15}} & {\textbf{R@30}} & {\textbf{SR@5}} & {\textbf{SR@15}} & {\textbf{SR@30}} & {\textbf{Rot.}} & {\textbf{Trans.}} \\
\specialrule{1.2pt}{0pt}{0pt}
\multirow{4}{*}{$\boldsymbol{\gamma_2}$}
& \textbf{0} & 0.593 & 0.864 & 0.884 & 0.562 & \textbf{0.818}  & 0.835 & 4.11 & 0.049 \\
& \textbf{0.1} & 0.629 & 0.866 & 0.886 & \textbf{0.582} & 0.808 & \textbf{0.836} & 3.61 & 0.048 \\
& \textbf{0.3} & 0.645 & \textbf{0.907} & \textbf{0.927} & 0.501 & 0.728 & 0.755 & 3.57 & \textbf{0.045} \\
& \textbf{0.5} & \textbf{0.677} & \textbf{0.907} & 0.923 & 0.395 & 0.582 & 0.612 & \textbf{3.41} & 0.048 \\
\specialrule{1.5pt}{0pt}{0pt}
\end{tabular}
\end{adjustbox}
\label{tab:noise_scale}
\vspace{-0.5em}
\end{table}
\noindent \textbf{Ablation on Noise Scale $\gamma$.} This ablation demonstrates that the noise scale $\gamma$ in Eq.~(\ref{eq:update_function}) provides a mechanism to trade off between recall and success rate. As shown in Table~\ref{tab:noise_scale}, increasing the noise scale leads to higher recall but decreases success rate. This is because a larger noise scale increases the variance of the first stage predictions, as indicated in Eq.~(\ref{eq:decay_converge}), allowing the second-stage energy-based optimization to start from more diverse initial points. Some of these diverse points may lie closer to the global optimum, thereby improving recall. However, others may start farther from feasible or optimal regions, which can reduce the success rate. Overall, the noise scale $\gamma$ effectively controls the balance between exploration (recall) and convergence stability (success rate). 
In our experiments, we also found that adding noise to the latitude and radius coordinates significantly decreases recall. We conjecture that this is because the energy landscape is steeper along these dimensions, so even small perturbations in latitude or radius can cause large changes in energy, resulting in implausible starting points for second stage refinement.

\begin{table}[t]
\centering
\caption{\textbf{Ablation on pose-conditioned priors}. Comparison of different pose-conditioned diffusion models used for second-stage refinement.}
\vspace{-0.5em}
\renewcommand{\arraystretch}{1.2}
\begin{adjustbox}{width=\linewidth}
\begin{tabular}{l c c c c c c c c}
\specialrule{1.5pt}{0pt}{0pt}
& {\textbf{R@5}} & {\textbf{R@15}} & {\textbf{R@30}} & {\textbf{SR@5}} & {\textbf{SR@15}} & {\textbf{SR@30}} & {\textbf{Rot.}} & {\textbf{Trans.}} \\
\specialrule{1.2pt}{0pt}{0pt}
\textbf{Zero123-XL} & \textbf{0.645} & \textbf{0.907} & \textbf{0.927} & \textbf{0.501} & \textbf{0.728}  & \textbf{0.755} & \textbf{3.57} & \textbf{0.045} \\
\textbf{Zero123} & 0.534 & 0.880 & 0.904 & 0.389 & 0.659& 0.699 & 4.52 & 0.054 \\
\specialrule{1.5pt}{0pt}{0pt}
\end{tabular}
\end{adjustbox}
\label{tab:ablation_model}
\vspace{-1em}
\end{table}
\noindent\textbf{Ablation on Pose-Conditioned Priors.}\hspace{1em} 
We study different pose-conditioned diffusion models for second-stage refinement, including Zero123 and Zero123-XL. As shown in Table~\ref{tab:ablation_model}, using Zero123-XL consistently improves performance across all evaluation metrics. This improvement occurs because the refinement stage relies on energy-based optimization, whose landscape is implicitly defined by the pretrained pose-conditioned diffusion model. In other words, the optimization aims to determine the relative pose that best matches the query view. Consequently, a stronger generative prior naturally leads to a more effective and robust refinement process.

\begin{figure*}[t]
  \centering

  \begin{minipage}{\textwidth}
    \centering
    \includegraphics[width=\textwidth]{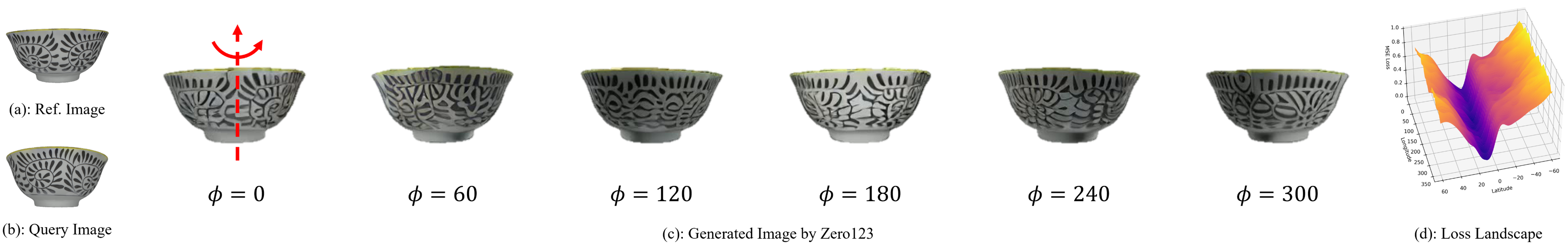}
    \caption{
    (a) Reference image and (b) query image of the object, captured from different camera poses. 
    (c) Image generated by Zero123 by varying $\phi$ from $0^{\circ}$ to $300^{\circ}$ in $60^{\circ}$ increments. 
    (d) Corresponding MSE loss landscape.}
    \label{fig:failure_case_bowl}
  \end{minipage}
  \hfill
  \begin{minipage}{\textwidth}
    \centering
    \includegraphics[width=\textwidth]{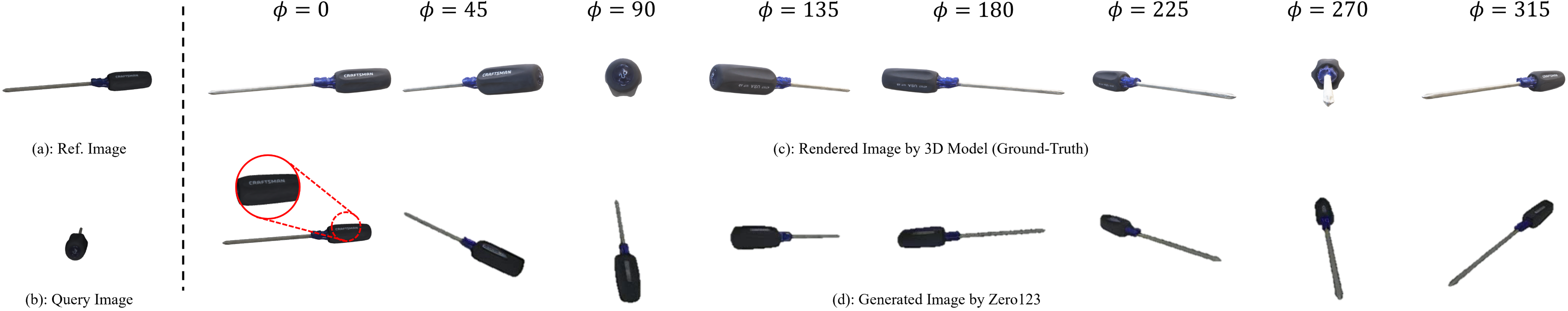}
    \caption{
    (a) Reference image and (b) query image of the object, captured from different camera poses. (c) Rendered image from the 3D CAD model. (d) Image generated by the Zero123 model. 
    Both vary $\phi$ from $0^{\circ}$ to $315^{\circ}$ in $45^{\circ}$ increments.}
    \label{fig:failure_case_screwdriver}
  \end{minipage}

\end{figure*}

\noindent \textbf{Failure Case Analysis.}\hspace{1em} While our method significantly improves the overall success rate and recall across most objects, there remain certain failure cases where performance degrades. To better understand these issues, we analyze two representative examples.

As shown in the Figs.~\ref{fig:failure_case_bowl}~(a) and (b), the Bowl object exhibits a continuous symmetrical shape along its axis and features repetitive surface patterns. Even for humans, estimating its horizontal rotation angle is challenging due to the lack of distinctive visual cues. To investigate whether the performance drop stems from limitations of Zero123, we visualize novel views of the object generated by Zero123, as shown in Fig.~\ref{fig:failure_case_bowl}~(c). The results indicate that Zero123 also struggles to generate meaningful and consistent views for this object, particularly when varying the longitude. Therefore, the refinement stage no longer provides any improvement in this case.

The second example involves the Screwdriver object, as shown in Fig.~\ref{fig:failure_case_screwdriver}. Unlike the previous case, this failure arises from the pose representation used by Zero123. Since Zero123 conditions on the difference in spherical coordinates between two views, it must implicitly estimate the pose of the object in the reference image to uniquely define the novel view's pose. However, in this example, the camera pose in reference image is positioned directly in front of the screwdriver, making it impossible to determine the object's latitude. To better understand this issue, we vary $\phi$ from $0^{\circ}$ to $315^\circ$ in $45^\circ$ increments and generate images using Zero123, as shown in Fig.~\ref{fig:failure_case_screwdriver}~(d). From these results, we hypothesize that Zero123 interprets the camera pose of the reference image as being from above the object. In reality, the ground-truth $\theta$ value is nearly $0^{\circ}$, indicating that the camera is positioned at the side. The ground-truth views for this object are shown in Fig.~\ref{fig:failure_case_screwdriver}~(c). 


\section{Visualization}
\label{sec:visualization}

In this section, additional visualizations are provided. Section~\ref{sec:additional_visualization_of_zero123_mse_loss_landscape} presents the Zero123 MSE loss landscape to supplement the analysis in the main manuscript. Section~\ref{sec:visualization_of_optimization_trajectories} illustrates the optimization trajectories. Section~\ref{sec:additional_qualitative_results} shows qualitative comparisons between iFusion and our method on camera pose estimation. Section~\ref{sec:visualization_of_score_field} visualizes the learned score and energy fields. Collectively, these results offer deeper insights into the optimization dynamics and demonstrate the effectiveness of our proposed approach.

\begin{figure*}[t]
  \centering
  \includegraphics[width=0.8\textwidth]{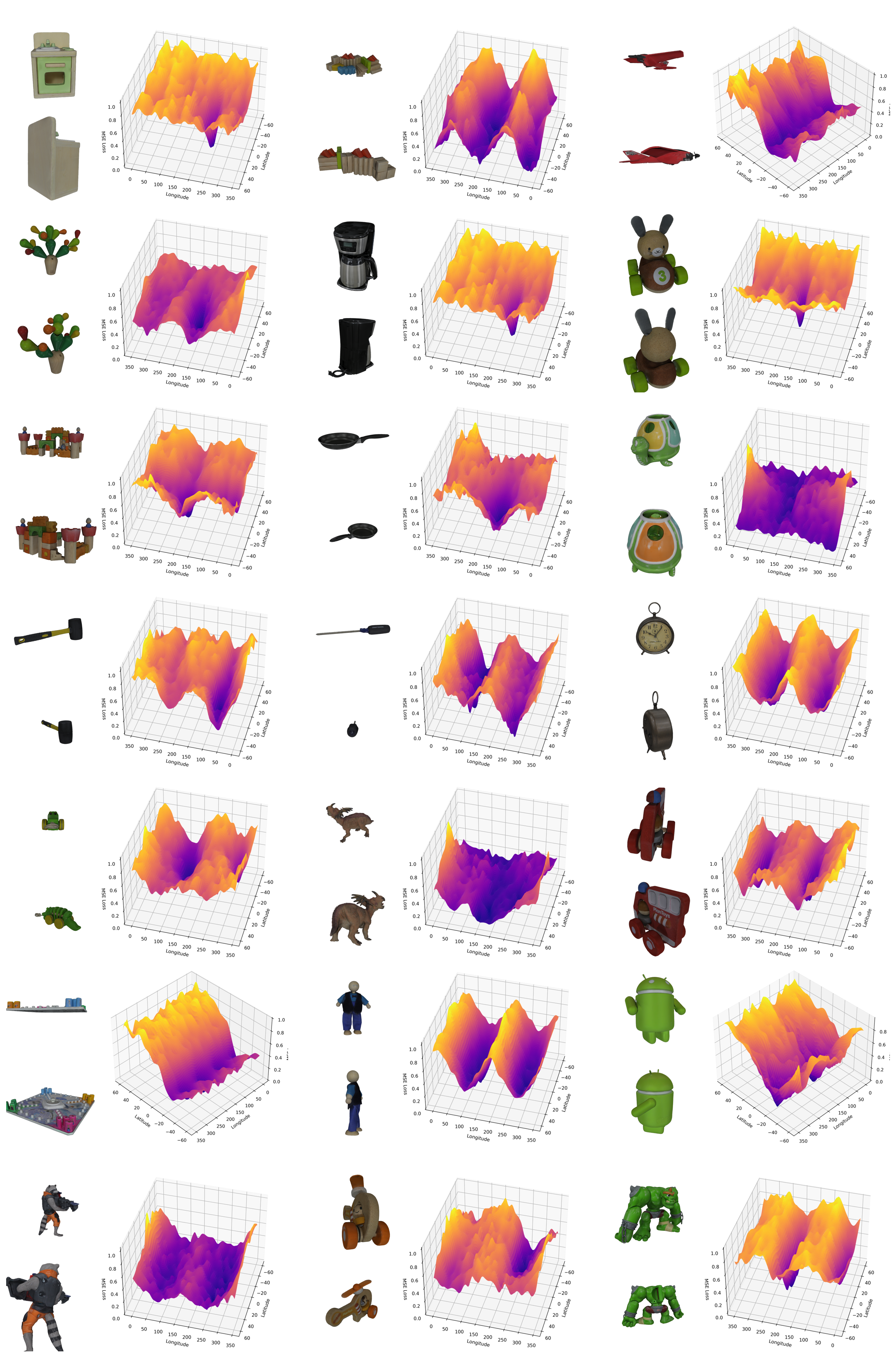}
  \caption{\textbf{3D MSE Landscape.} Using image pairs from the GSO dataset \citep{DBLP:conf/icra/DownsFKKHRMV22}, we visualize the Zero123 MSE loss landscape for each object. The plots clearly reveal the presence of local minima and plateau regions, highlighting inherent optimization challenges.}
  \label{fig:l2_3d_landscape}
\end{figure*}

\normalsize{\subsection{\textbf{Additional Zero123 MSE Landscape Visualization}}}
\label{sec:additional_visualization_of_zero123_mse_loss_landscape}
We provide further details on visualizing the MSE loss landscape with Zero123, as shown in Fig.~\ref{fig:ifusion_landscape}. Given a pair of images, we follow the Zero123 noise prediction pipeline illustrated in Fig.~\ref{fig:architecture}~(b) of the main manuscript to compute MSE loss. To explore the loss landscape, we sweep the conditioning pose over latitude from $-60^\circ$ to $60^\circ$ in steps of $8^\circ$ and longitude from $0^\circ$ to $360^\circ$ in steps of $12^\circ$. At each pose, the MSE loss is computed and averaged over five random noise samples. Additional examples are shown in Fig.~\ref{fig:l2_3d_landscape}, offering a more comprehensive visualization of the results. These extended visualizations further demonstrate that the Zero123 MSE loss landscape contains multiple local minima and flat regions (plateaus).

\subsection{Visualization of Optimization Trajectories}
\label{sec:visualization_of_optimization_trajectories}
\begin{figure*}[t]
  \centering
  \includegraphics[width=0.88\textwidth]{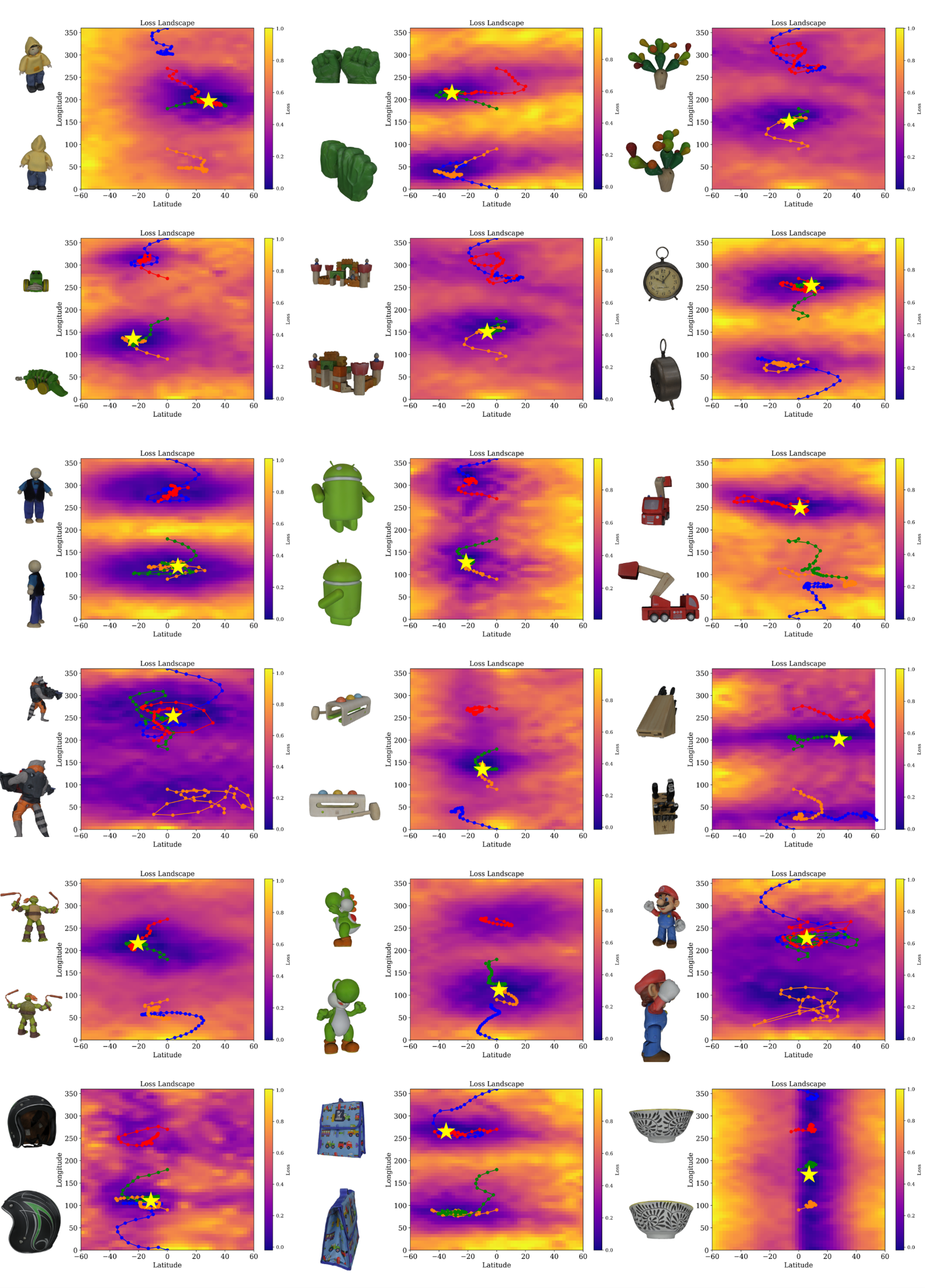}
  \caption{\textbf{Optimization Trajectories.} Trajectories generated by optimizing the conditioning pose with Zero123 using MSE loss. Starting from four initial longitudes ($0^\circ$, $90^\circ$, $180^\circ$, $270^\circ$), only some converge to the ground-truth pose, while others fall into local minima.}
  \label{fig:ifusion_trajs}
\end{figure*}

Additional examples corresponding to Fig.~\ref{fig:trajs}~(d) are presented in Fig.~\ref{fig:ifusion_trajs}. These visualization clearly demonstrate the local minimum issue that occurs when directly optimizing the conditioning pose using Zero123 with the MSE loss. For each case, the optimization process is initialized from four different starting poses at longitudes $0^\circ$, $90^\circ$, $180^\circ$, and $270^\circ$. In most cases, only one or two of the initial points converge to the ground-truth pose, while the others become trapped in local minima. 

\subsection{Additional Qualitative Results}
\label{sec:additional_qualitative_results}
\begin{figure*}[t]
  \centering
  \includegraphics[width=0.8\textwidth]{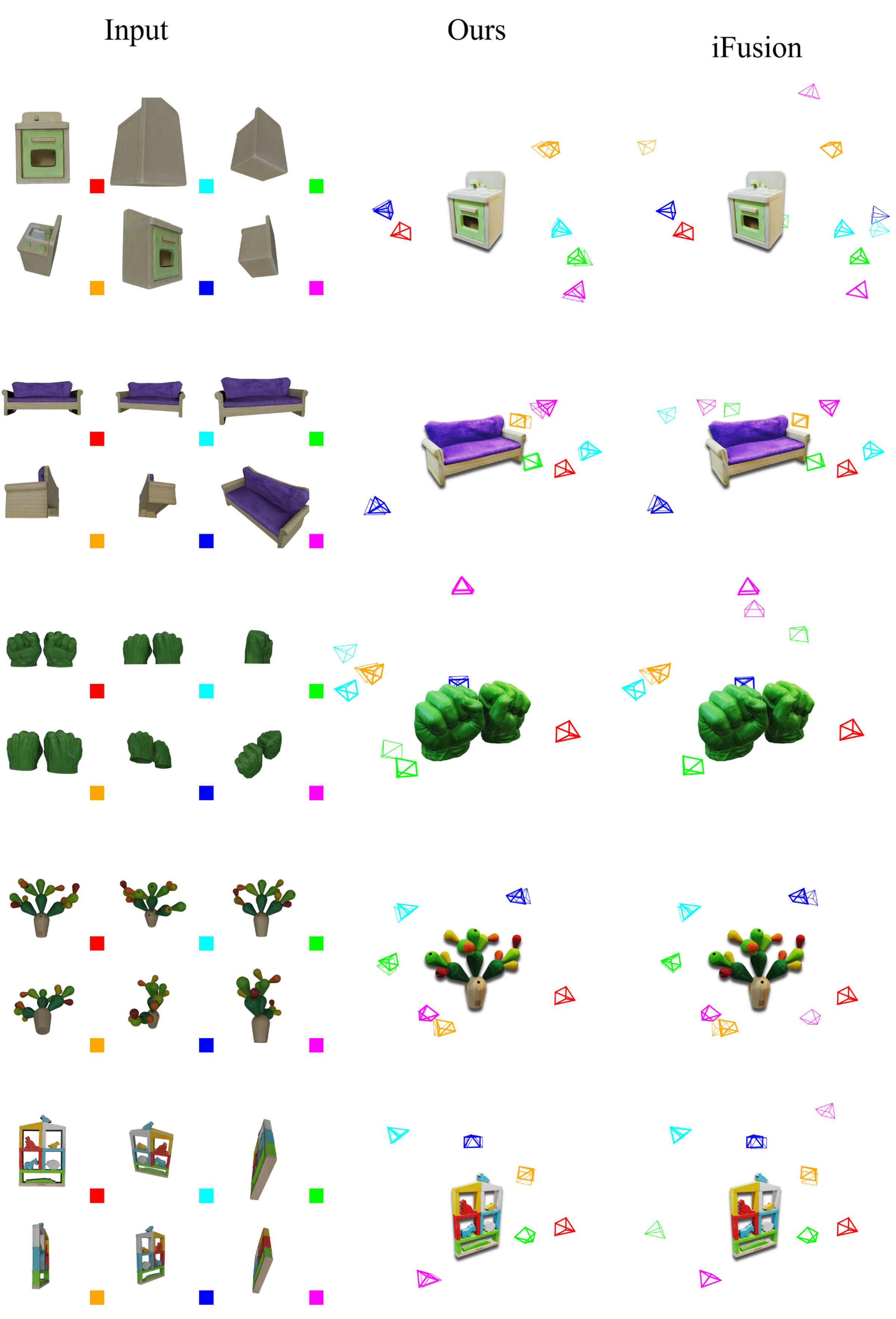}
  \caption{\textbf{Qualitative Results for Camera Pose Estimation.} The figure visualizes predicted camera poses (thin) alongside ground-truth poses (bold). For each object, we estimate the relative camera poses of five target views from a single reference image, shown in \textbf{\textcolor{red}{red}}. Both methods start from two randomly initialized poses. Our method consistently converges to the correct poses, while iFusion often gets stuck in local minima, resulting in incorrect predictions.}
  \label{fig:qualitative_results}
\end{figure*}

Additional qualitative results comparing iFusion and our method are shown in Fig.~\ref{fig:qualitative_results}. These results clearly demonstrate that our gradient-based two-stage optimization effectively avoids local minima and consistently guides the pose updates toward the ground-truth pose. Importantly, our method is robust to different initial points and achieves strong performance across a wide range of starting poses, on par with state-of-the-art approaches.

\subsection{Visualization of Score Field}
\label{sec:visualization_of_score_field}
\begin{figure*}[t]
  \centering
  \includegraphics[width=0.78\textwidth]{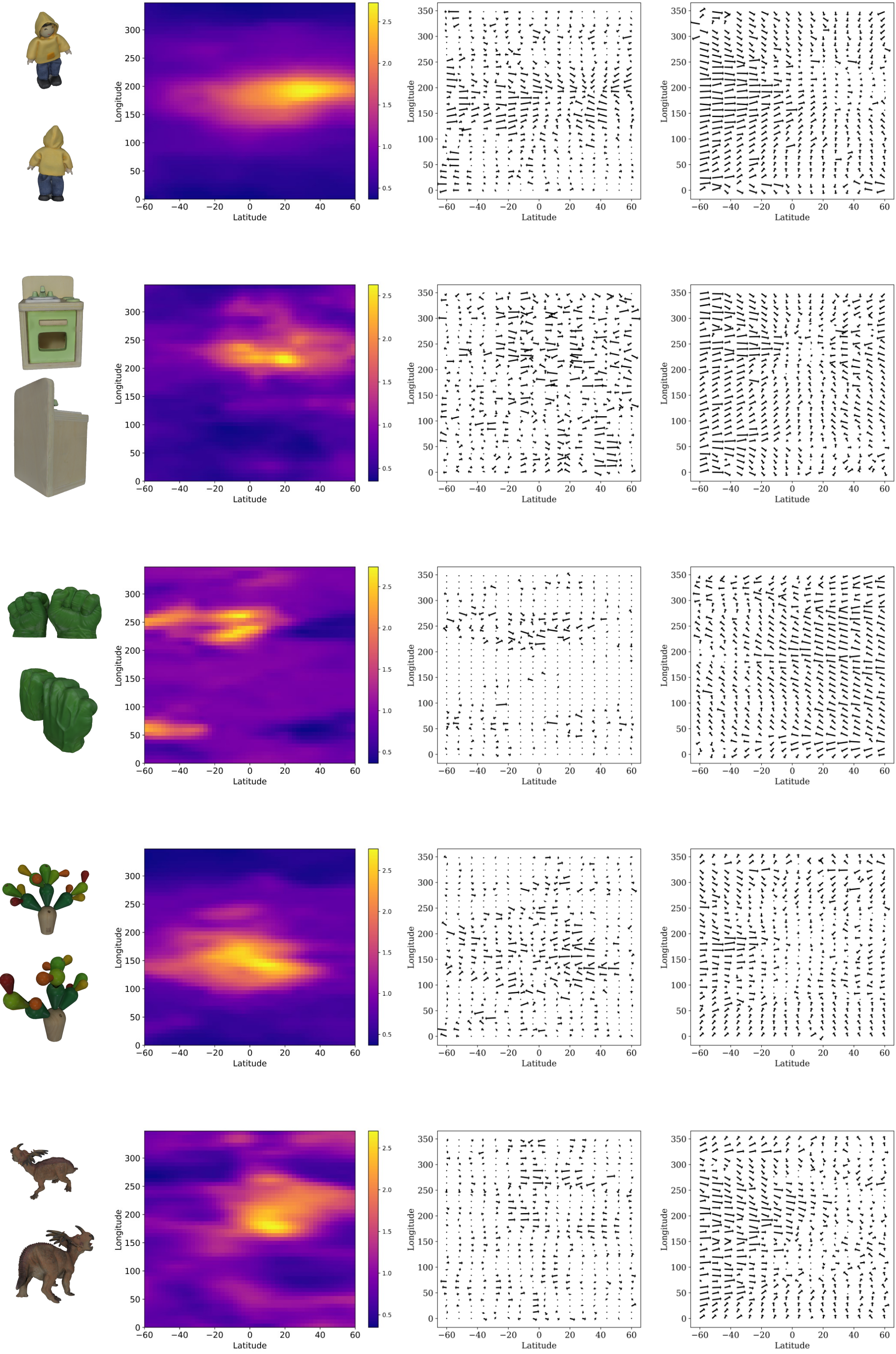}
  \caption{\textbf{Comparison Between Score and Energy.} The leftmost column shows the image pair, with the top image as the reference and the bottom image as the query. The second column shows the learned energy field, visualized as $\exp (-\mathcal{E}(x))$, where $\mathcal{E}(x)$ denotes the energy function. The higher value represents the high probability region. The third column shows the score field corresponding to the energy field, calculated by automatic differentiation. The last column shows the score field learned by the score-based method.}
  \label{fig:energy_score_comparison}
\end{figure*}

In Fig.~\ref{fig:energy_score_comparison}, we visualize the score and energy fields from the energy-based method, alongside the score field from the score-based method. The energy-based model obtains the score by differentiating the learned energy, while the score-based model directly learns it via score matching, resulting in a consistently more accurate score field.

\section{Limitations}
\label{sec:limitations}
From the failure case analysis in Section~\ref{sec:more_ablation_studies}, we identify two limitations of our method. 
The first limitation arises from the limited generative capability of Zero123 model on specific objects. Due to multi-view inconsistencies in Zero123, refinement can be challenging for such objects. Some recent works~\citep{chen2024cascade, shi2023zero123++} address this issue and show promising improvements. Combing these pose-conditioned diffusion models with our score-based guidance framework suggests a promising direction for future work.
The second limitation concerns the pose representation used in Zero123. The model represent relative poses in polar coordinates without explicitly defining an object coordinate system. This leads to ambiguity, especially for symmetric objects. Consequently, the same relative pose may correspond to multiple plausible target views, making pose-conditioned generation inherently ambiguous and adversely affecting our method’s performance.
We leave these two limitations as opportunities for future improvement.


\end{document}